\documentclass[twoside]{article}

\usepackage{booktabs}
\usepackage{multirow, makecell}
\usepackage{arydshln}
\usepackage{thmtools} 
\usepackage{thm-restate}
\usepackage{amsthm}
\usepackage{mathtools}
\usepackage{bbold}

\usepackage[labelfont=bf]{caption}
\usepackage{subcaption}

\usepackage{url}
\usepackage[linesnumbered,lined,ruled]{algorithm2e}

\usepackage{titletoc}
\usepackage{tocloft}
\usepackage{lipsum}

\usepackage[accepted]{aistats2024}

\usepackage[round]{natbib}

\usepackage[	
    colorlinks=true,
	linkcolor=black,
	citecolor=black,
    urlcolor=black,
	]{hyperref}  
\usepackage[nameinlink]{cleveref}

\usepackage{tikz}
\usepackage{tikz-3dplot} 
\usepackage{pgfplots}

\usepackage{macro}

\newcommand{\jake}[1]{}
\newcommand{\kw}[1]{}
\newcommand{\geoff}[1]{}
\newcommand{\jw}[1]{}

\newcommand{\papertitle}{Large-Scale Gaussian Processes via Alternating Projection}

\begin{document}

\twocolumn[
\aistatstitle{\papertitle}
\runningauthor{Wu, Wenger, Jones, Pleiss, Gardner}
\aistatsauthor{Kaiwen Wu$^1$ \quad Jonathan Wenger$^2$ \quad Haydn Jones$^1$ \quad Geoff Pleiss$^{3,4}$ \quad Jacob R. Gardner$^1$}
\vspace{0.2em}
\aistatsaddress{$^1$University of Pennsylvania \quad $^2$Columbia University \quad $^3$University of British Columbia \quad $^4$Vector Institute} 
]

\begin{abstract}
Training and inference in Gaussian processes (GPs) require solving linear systems with $n\times n$ kernel matrices. To address the prohibitive $\mathcal{O}(n^3)$ time complexity, recent work has employed fast iterative methods, like conjugate gradients (CG). However, as datasets increase in magnitude, the kernel matrices become increasingly ill-conditioned and still require $\mathcal{O}(n^2)$ space without partitioning. Thus, while CG increases the size of datasets GPs can be trained on, modern datasets reach scales beyond its applicability. In this work, we propose an iterative method which only accesses subblocks of the kernel matrix, effectively enabling mini-batching. Our algorithm, based on alternating projection, has $\mathcal{O}(n)$ per-iteration time and space complexity, solving many of the practical challenges of scaling GPs to very large datasets. Theoretically, we prove the method enjoys linear convergence. Empirically, we demonstrate its fast convergence in practice and robustness to ill-conditioning. On large-scale benchmark datasets with up to four million data points, our approach accelerates GP training and inference by speed-up factors up to $27\times$ and $72 \times$, respectively, compared to CG.
\end{abstract}

\section{INTRODUCTION}
\begin{figure}[t]
\centering
\begin{subfigure}[t]{0.49\linewidth}
    \includegraphics[width=\textwidth]{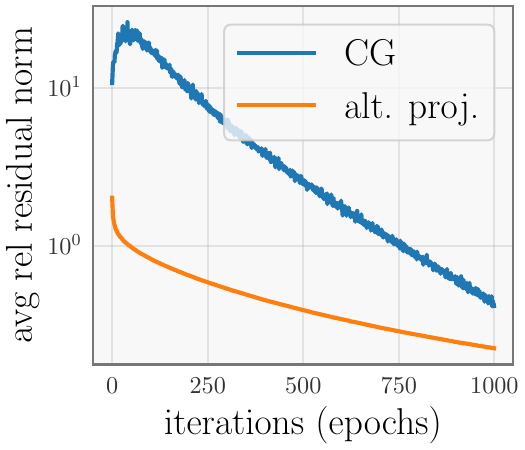}
    \caption{3droad}
\end{subfigure}
\begin{subfigure}[t]{0.49\linewidth}
    \includegraphics[width=\textwidth]{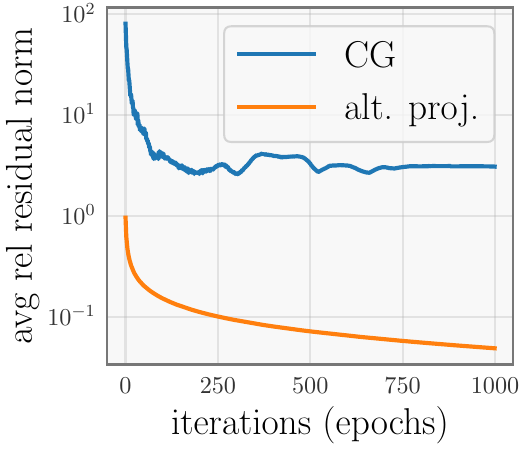}
    \caption{house electric}
\end{subfigure}
\caption{
Convergence of alternating projection and (preconditioned) conjugate gradient.
The $x$-axis is the number iterations for CG and the number epochs for alternating projection.
Both methods are initialized at zero, but CG increases the residual after the first iteration.
\textbf{Left:} While the asymptotic convergence rate of CG can be faster than alternating projection, CG does not find a better solution than alternating projection in the first $1000$ iterations.
\textbf{Right:} CG struggles with convergence due to ill-conditioning and does not reach the tolerance $\epsilon = 1$. In contrast, alternating projection convergences.
See \S\ref{sec:convergence} for more details.
}
\label{fig:first-figure}
\end{figure}

Scaling Gaussian process (GP) models to large datasets has been a central research topic in probabilistic machine learning for nearly two decades. The primary challenge is the cubic complexity of computing both the marginal log likelihood (MLL) during training and the predictive distribution at test time.
Over the years, this problem has been addressed both from a modeling perspective \citep[\eg,][]{hensman2013gaussian,hensman2015scalable,titsias2009variational,snelson2005sparse,salimbeni2018orthogonally,jankowiak2020parametric,katzfuss2021general}
and from a numerical methods perspective \citep[\eg,][]{cutajar2016preconditioning,pleiss2018constant,gardner2018gpytorch,wang2019exact,maddox2022low},
and contemporary work even unifies these perspectives to a degree \citep{artemev2021tighter,wenger2022posterior}.
In recent years, numerical methods have increasingly relied on matrix-free iterative methods, which access the kernel matrix through matrix-vector multiplications.
These iterations are suitable for GPU acceleration \citep{gardner2018gpytorch} and have shown success on medium to moderately large datasets \citep{wang2019exact}, outperforming modeling-based approaches such as stochastic variational GPs (SVGP) \citep{hensman2013gaussian}.

Most GP training and inference approaches based on iterative methods use classic general-purpose algorithms for matrix solves,
such as conjugate gradients (CG) \citep{cutajar2016preconditioning,gardner2018gpytorch,wang2019exact},
MINRES \citep{pleiss2020fast},
or (stochastic) gradient descent \citep{lin2023sampling}.
There is reason to believe that such algorithms are suboptimal for modern hardware-accelerated Gaussian processes.
For example, CG was purpose-built for sparse linear systems that require high-precision solutions.
Neither of these properties applies to GP regression:
the necessary solves involve dense covariance matrices,
and tasks such as hyperparameter optimization can be performed with extremely coarse-grained solves \citep{wang2019exact,maddox2022low}.
These characteristics of large-scale dense operations and low precision amenability are in line with existing trends in machine learning
\citep{courbariaux2015binary,micikevicius2018mixed},
but ultimately place Gaussian processes at odds with much of the literature on numerical methods.

Much in the way that deep learning has been revolutionized by purpose-built optimizers that exploit properties of neural networks \citep{kingma2015adam,loshchilov2017decoupled},
this paper aims to accelerate GPs with a purpose-built method leveraging (coarse-grained) covariance matrix solves on modern hardware.
We introduce an iterative method to compute gradients of the marginal log-likelihood (MLL) and the posterior mean, that improves over CG in the following ways:
1) it requires $\Oc(n)$ computation per iteration rather than CG's $\Oc(n^2)$;
2) it converges rapidly and monotonically in its early stages (but does not necessarily obtain higher precision than CG); and
3) it demonstrates improved numerical stability in floating point arithmetic.

\paragraph{Contributions.}
We propose an iterative method for Gaussian process training and inference.
The method computes the marginal log-likelihood derivative and posterior mean via alternating projection.
Each iteration of the algorithm accesses a subblock of the kernel matrix, has linear time and memory complexity, and decreases the residual near-monotonically after every epoch.
We prove that the algorithm converges linearly at a rate no slower than gradient descent, despite never operating on the full kernel matrix.
Empirically, our method achieves a speed-up of up to $27\times$ over CG-based hyperparameter training and of up to $72 \times$ over CG-based inference on a wide range of datasets.
As a demonstration of its scalability and robustness to ill-conditioning, we are able to train Gaussian processes on $4$ million data points, the largest dataset reported in the literature to-date without using inducing points or similar modeling approximations---to the best of our knowledge. We find that our method outperforms the stochastic variational Gaussian process by a significant margin at this scale.

\section{SETUP AND BACKGROUND}
\label{sec:background}

\paragraph{Notation.}
Let \((\Xv, \yv)\) be a training set of \(n\) training inputs
$\Xv = (\begin{matrix}\xv_1 & \cdots & \xv_n\end{matrix})^\top \in \mathcal X \subseteq \Rb^{n \times d}$
and labels $\yv = (\begin{matrix}y_1 & \cdots & y_n\end{matrix})^\top \in \Rb^{n}$.
Let the set $\{1, 2, \dots, n\}$ be denoted by $[n]$.
Given a matrix $\Av \in \Rb^{n \times n}$
and an index set $I \subseteq [n]$,
$\Av_I = \Av_{I, :}$ is the $|I| \times n$ row-indexed submatrix,
$\Av_{:, I}$ the $n \times |I|$ column-indexed submatrix,
and $\Av_{I, I}$ is the $|I| \times |I|$ principal submatrix.
We use similar indexing notations for vectors.

Let $\Ev \in \Rb^{n \times n}$ be the identity matrix.
$\Ev_I$ denotes the $\lvert I \rvert \times n$ submatrix formed by rows indexed by $I$.
Notice that multiplication with $\Ev_I \in \Rb^{|I| \times n}$ selects rows and columns: $\Ev_I \Av = \Av_{I}$ and $\Av \Ev_I^\top = \Av_{:, I}$ for any $n \times n$ matrix $\Av$.
For a vector $\uv \in \Rb^{\lvert I \rvert}$, left multiplying $\Ev_I^\top$ maps $\uv$ to a $n$-dimensional vector $\vv$, such that $\vv_I = \uv$ and the entries outside $I$ are zeros: $\vv_{[n] \setminus I} = 0$, where $[n] \setminus I$ is the complement of $I$.

Now, let $f: \Xc \to \Rb$ be a latent function,
and let
$k_{\thetav}: \Xc \!\times\! \Xc \to \Rb$ be a (known) positive definite kernel function
with hyperparameters $\thetav$.
We write $\fv=f(\Xv)=(\begin{matrix}f(\xv_1) & \cdots & f(\xv_n)\end{matrix})^\top \in \Rb^n$.
Similarly, $k_{\thetav}(\Xv, \cdot): \Xc \to \Rb^n$ denotes the vector-valued function given by $(\begin{matrix}k(\xv_1, \cdot) & \cdots & k(\xv_n, \cdot)\end{matrix})^\top$,
and $\Kv_{\thetav} \in \Rb^{n \times n}$ is the Gram matrix with $[\Kv_{\thetav}]_{ij} = k_{\thetav}(\xv_i, \xv_j)$.
We omit the subscript $\thetav$ unless the context needs it.

\paragraph{Gaussian Process Regression.}
In supervised GP regression, we assume a response-generating function $f$ that is Gaussian process distributed a priori---\ie $f \sim \mathcal{GP} \big(\mu, k_{\thetav})$.
For simplicity of presentation, we assume without loss of generality an exact observation model---\ie $\yv = f(\Xv)$.\footnote{Note that we can easily recover an observational noise model by setting $k_{\thetav}(\xv, \xv^\prime) = k_\mathrm{base}(\xv, \xv^\prime) + \sigma^2 \mathbb{1}[\xv=\xv^\prime]$ for some $k_\mathrm{base}$ and $\sigma > 0$, where $\mathbb{1}$ is the indicator function.}
Given a finite test dataset $\xv_1^*, \ldots, \xv_M^*$, we can obtain a posterior distribution over $f(\xv^*_1), \ldots, f(\xv^*_M)$
using standard Gaussian conditioning rules with the posterior mean and covariance:
\begin{align*}
\Eb[\fv^* \mid \fv] & = \muv + \Kv_{*\fv} \Kv^{-1} (\yv - \muv), \\
\Db[ \fv^* \mid \fv] & = \Kv_{**} - \Kv_{*\fv} \Kv^{-1} \Kv_{\fv*}.
 \end{align*}
We refer the reader to \citet[][Ch.~2]{rasmussen2006gaussian} for more details.

\paragraph{Hyperparameter Training.}
The hyperparameters $\thetav$ of the GP are learned by minimizing the negative marginal log likelihood (MLL)
$\ell(\thetav) \vcentcolon= -\log p(\yv ; \thetav)$.
With a Gaussian process prior on $f$, we have $p(\yv; \thetav) = \mathcal{N}(\yv; \muv, \Kv_{\thetav})$,
yielding the following minimization:
\begin{align}
    \mini_{\thetav} \ell(\thetav) \overset{c}{=} {\textstyle \frac{1}{2}}\left(\yv^\top \Kv_{\thetav}^{-1} \yv + \log\det(\Kv_{\thetav})\right)
    \label{eqn:loss}
\end{align}
\Cref{eqn:loss} is commonly optimized with first-order methods,
which require an (unbiased) estimate of $\frac{\partial \ell(\thetav)}{\partial \theta}$. Unfortunately, as \eqref{eqn:loss} cannot be written in the usual $\sum_{i=1}^{n}\ell(\xv_i, y_i)$ form common to many machine learning algorithms, standard minibatching strategies are not readily applicable.
Following prior work \citep[\eg][]{cutajar2016preconditioning,gardner2018gpytorch,wenger2022preconditioning},
we use the following unbiased estimate:
\begin{align}
    - \tfrac 1 2 \yv^\top \Kv_{\thetav}^{-1} \tfrac{\partial \Kv_{\thetav}}{\partial \theta} \Kv_{\thetav}^{-1} \yv + {\textstyle \frac{1}{2l} \sum_{i = 1}^{l}}
    \Big(\zv_i^\top \Kv_{\thetav}^{-1}\Big) \tfrac{\partial \Kv_{\thetav}}{\partial \theta} \zv_i,
    \label{eqn:loss_grad_unbiased}
\end{align}
where $\zv_i$ are \iid stochastic trace samples with zero mean $\ep{\zv_i} = \zero$ and identity covariance $\ep{\zv_i \zv_i^\top} = \Ev$.
Note that the second term is an unbiased stochastic approximation of $\tr\bb[\big]{\Kv_{\thetav}^{-1} \frac{\partial \Kv_{\thetav}}{\partial \theta}}$.
Crucially, computing \eqref{eqn:loss_grad_unbiased} primarily involves computing linear solves with $\Kv_{\thetav}$.

\paragraph{Linear Solves with Iterative Methods.}
When the size of $\Kv$ is large, direct methods solving $\Kv \wv = \bv$ are prohibitively slow.
Iterative methods, such as conjugate gradients (CG),
offer reduced asymptotic complexity \citep{cutajar2016preconditioning},
significant GPU acceleration \citep{gardner2018gpytorch},
and memory savings if the kernel matrix $\Kv$ is accessed in a map-reduce fashion \citep{wang2019exact,charlier2021kernel}.

\jw{Mention somewhere that for hyperopt we need to solve the linear system \(\Kv \Wv = \begin{bmatrix} \yv & \zv_1 & \dots & \zv_l \end{bmatrix}=\Bv\)}

\kw{added below}

CG minimizes the quadratic objective
$\tfrac12 \wv^\top \Kv \wv - \bv^\top \wv$
by iteratively searching along conjugated directions.
Each iteration requires a $\Oc(n^2)$ matrix-vector multiplication with $\Kv$.
In exact arithmetic, CG returns an exact solution after $n$ iterations.
In practice for ill-conditioned problems, CG is terminated once the residual $\rv = \bv - \Kv \wv$ is small enough, \eg, $\| \rv \| \!\leq \epsilon \| \bv \|$ for some predefined tolerance parameter $\epsilon$.

For GP hyperparameter learning often large values of the tolerance $\epsilon$ are used despite the potential for overfitting \citep{potapczynski2021bias}.
For instance, $\epsilon = 1$ is used in practice \citep{wang2019exact,maddox2022low} and has been the default tolerance of CG during training in popular GP software packages, including GPyTorch\footnote{GPyTorch setting \url{https://rb.gy/qi8er}} and GPflow\footnote{GPflow setting \url{https://rb.gy/mozif}}.

For hyperparameter training, each MLL derivative evaluation requires a batched linear solve $\Kv \Wv = \Bv$, where $\Bv =  (\begin{matrix} \yv & \zv_1 & \dots & \zv_l \end{matrix})$ with $\zv_i$ are random samples for stochastic MLL derivative estimation in \eqref{eqn:loss_grad_unbiased}.

\paragraph{RKHS.}
Every kernel $k: \Xc \times \Xc \to \Rb$ induces a function space $\Hc = \overline{\mathrm{span}} \{k(\xv, \cdot): \xv \in \Xc\} \subseteq \Rb^{\Xc}$, known as a reproducing kernel Hilbert space (RKHS) 
where its inner product $\langle \cdot, \cdot \rangle_{\Hc}$ satisfies $\langle k(\xv, \cdot), k(\xv', \cdot) \rangle_{\Hc} = k(\xv, \xv')$ for all $\xv, \xv' \in \Xc$.

\paragraph{RKHS Projection.}
Given a set of indices $I \subseteq [n]$, define the finite dimensional linear subspaces of $\Hc$:
\begin{equation}
    \begin{aligned}
        V_{[n]} &\vcentcolon= \mathrm{span} \{k(\xv_i, \cdot): i = 1, 2, \cdots, n\} \subseteq \Hc,
        \\
        V_{I} &\vcentcolon= \mathrm{span} \{k(\xv_i, \cdot): i \in I\} \subseteq V_{[n]},
    \end{aligned}
    \label{eqn:subspaces}
\end{equation}
By definition these subspaces contain functions of the form $f(\cdot) = \sum_{i=1}^{n}\alpha_{i}k(\xv_i, \cdot)$ and $f(\cdot) = \sum_{i \in I}\alpha_{i}k(\xv_i, \cdot)$ respectively.
We can map any $f \in \Hc$ onto these subspaces using the projection operator.
\begin{definition}[Projection Operator]
\label{def:projection_operator}
Let $V \subseteq \Hc$ be a closed linear subspace.
The projection of any $f \in \Hc$ onto $V$ is
given by the \emph{projection operator}
\begin{align*}
    \proj_{V}(f) = \argmin_{g \in V} ~~ \tfrac12 \| f - g \|_{\Hc}^2,
\end{align*}
which is well-defined by the Hilbert space projection theorem, \ie, the unique minimizer exists.
\end{definition}
Intuitively, the projection operator finds the best approximation of $f$ inside $V$, where the approximation error is measured by the norm $\lVert \cdot \rVert_\Hc$.
For $V = V_{[n]}$ and $V = V_{I}$, the projection operator has a simple form:
\begin{equation}
    \begin{aligned}
    \label{eqn:proj_vn}
        \proj_{V_{[n]}} (f) &= f(\Xv)^\top \Kv^{-1} k(\Xv, \cdot), \\
        \proj_{V_I} (f) &= f(\Xv)^\top \Ev_{I}^\top \Kv_{I, I}^{-1} \Ev_{I} k(\Xv, \cdot).
    \end{aligned}
\end{equation}
Importantly, these projections only evaluate $f$ and the kernel $k$ on the training data $\Xv$ (or subset $\Xv_{I}$).
In other words, it is unnecessary to evaluate $f$ or $k$ outside of $\Xv$ (or $\Xv_{I}$).
The complexity of computing the projection $\proj_{V} (f)$ depends on the dimension of the subspace $V$.
A projection to $V_{[n]}$ takes $\Oc(n^3)$ time and a projection to $V_I$ takes $\Oc(\lvert I \rvert^3)$ time.

\section{METHOD}
\label{sec:method}

In this section, we develop an iterative method for computing solves $\Kv^{-1} \bv$ by alternating projection.
The method supports batch linear solves with multiple right-hand sides, as required by estimating the marginal log-likelihood (MLL) derivative \eqref{eqn:loss_grad_unbiased}, and is amenable to GPU parallelism.\jw{We need more explanation why this is necessary. Alternating projection is well-known. What do we develop here that's new? GPU specific method, tailored towards GPs, \dots}\kw{Added a few coments on the batch linear solve and amenable to GPU implementation}
We cast the linear solve as a projection in the RHKS $\Hc$ and decompose the projection into a sequence of small-scale subproblems.
Each subproblem is solved in $\Oc(n)$ time, allowing frequent updates.
An appealing feature of alternating projection, as we will see later later, is that it typically makes rapid progress in the early stage and finds a medium-precision solution quickly, which are already good enough for GP training and predictions.

\begin{figure}
\centering
\begin{subfigure}{0.48\linewidth}
\tdplotsetmaincoords{60}{120} 

\begin{tikzpicture} [scale=2.9, tdplot_main_coords, axis/.style={->,black,thick}, 
vector/.style={-stealth,red,very thick}, 
vector guide/.style={dashed,black,thick}]

\coordinate (O) at (0,0,0);

\pgfmathsetmacro{\ax}{0.3}
\pgfmathsetmacro{\ay}{0.7}
\pgfmathsetmacro{\az}{0.45}

\coordinate (P) at (\ax,\ay,\az);
\node at (P) [right] {\scriptsize $g = r^{(0)}$};

\coordinate (s) at (\ax,\ay,0);
\coordinate (r) at (0,0,\az);

\coordinate (x1) at (.5, 0, 0);
\coordinate (x2) at (0, .85, 0);
\coordinate (x3) at (0, 0, .7);

\draw[fill=blue!10] (O) -- (0,0.7,0) -- (0.3,0.7,0) -- (0.3,0,0) -- cycle;

\draw[axis] (O) -- (x1) node[anchor=north]{{\scriptsize $k(\xv_1, \cdot)$}};
\draw[axis] (O) -- (x2) node[anchor=north]{{\scriptsize $k(\xv_2, \cdot)$}};
\draw[axis] (O) -- (x3) node[anchor=north west]{{\scriptsize $k(\xv_3, \cdot)$}};

\draw[vector] (O) -- (P);

\draw[vector guide] (\ax,\ay,0) -- (P);
\draw[vector guide]         (P) -- (0,0,\az);

\draw[vector,blue] (O) -- (s);
\node at (s) [below left] {{\scriptsize $s_1$}};

\node at (r) [left] {{\scriptsize $r^{(1)}$}};
\draw[vector,blue] (O) -- (r);

\end{tikzpicture}
\end{subfigure}
\begin{subfigure}{0.48\linewidth}
\centering
    \includegraphics[width=\linewidth]{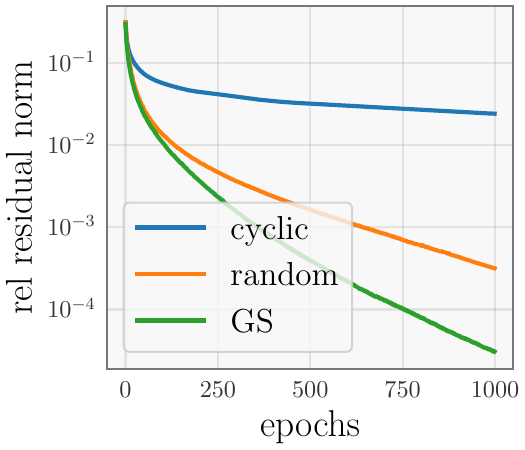}    
\end{subfigure}
\caption{
\textbf{Left:} Illustration of alternating projection.
$s^{(1)}$ is the projection of $g = r^{(0)}$ onto the subspace spanned by $k(\xv_1, \cdot)$ and $k(\xv_2, \cdot)$.
The residual $r^{(1)} = g - s^{(1)}$ will be projected to other coordinates in the next iteration.
\textbf{Right:} Gauss-Southwell block selection rule results in faster convergence than random and cyclic selection rules.
}
\label{fig:illustration_alternating_projection}
\end{figure}
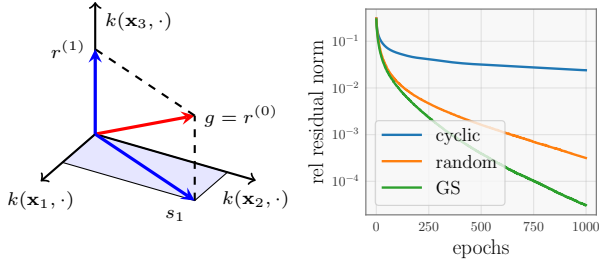

\paragraph{High Level Approach.}
Let $k$ be strictly positive definite and assume there is no duplicate data.
Then there exists $g \in \Hc$ interpolating $\bv$, \ie, $g(\Xv) = \bv$.
The exact form of $g$ is not important (or unique for that matter); rather, we are interested in its projection onto the subspace $V_{[n]}$, which by
\eqref{eqn:proj_vn} is
\begin{align*}
    \proj_{V_{[n]}}(g) = \bv^\top \Kv^{-1} k(\Xv, \cdot).
\end{align*}
Thus, the solution $\Kv^{-1} \bv$ can be obtained from the coefficients of the projection $\proj_{V_{[n]}}(g)$.
At a first glance, it is not even clear how to come up with this function $g$, let alone computing its projection.
As we will see soon, we do not need an explicit representation of $g$.

Directly projecting $g$ onto $V_{[n]}$ is computationally infeasible, as the time complexity is cubic in $n$.
Instead, we partition $[n]$ into subsets $\Pc = \{I_1, I_2, \cdots, I_m\}$.
For each subset $I \in \Pc$, the projection to the linear subspace $V_I \subseteq V_{[n]}$ is cheap, provided that $\lvert I \rvert$ is small.
Thus, we construct the (full) projection $\proj_{V_{[n]}} (g)$ by iteratively computing the projection onto the linear subspaces $V_I$ where $I \in \Pc$.

Starting from $r^{(0)} = g$ and $s^{(0)} = 0$, the $j$-th iteration selects an index set $I \subseteq [n]$ and updates as follows
\begin{align}
\label{eq:update_s}
s^{(j + 1)} = s^{(j)} + \proj_{V_I} \bb[\big]{r^{(j)}} \\
\label{eq:update_r}
r^{(j + 1)} = r^{(j)} - \proj_{V_I} \bb[\big]{r^{(j)}}
\end{align}
Intuitively, $s^{(j)}$ progressively approximates the true projection $\proj_{V_{[n]}} (g)$, since \eqref{eq:update_s} iteratively adds the projection onto subspaces $V_I$ to the current approximation $s^{(j)}$.
Meanwhile, \eqref{eq:update_r} accordingly updates the residual---the difference bewteen $g$ and $s^{(j)}$.
As $j \to \infty$, $s^{(j)}$ converges to the true projection $\bv^\top \Kv^{-1} k(\Xv, \cdot)$ \citep{wendland2004scattered}.
See \Cref{fig:illustration_alternating_projection} (left panel) for an illustration of alternating projection.

\paragraph{Store $r^{(j)}$ Implicitly.}
Crucially, in the updates \eqref{eq:update_s} and \eqref{eq:update_r},
the function $r^{(j)}$ is only ever accessed through its evaluation on the training data $\Xv$ (recall the projection formula \eqref{eqn:proj_vn}).
Therefore, we only need to maintain the residual vector $\rv^{(j)} = r^{(j)}(\Xv) \in \Rb^n$ instead of the entire function.
The update \eqref{eq:update_r} thus reduces to
\begin{align}
\rv^{(j + 1)} &= \rv^{(j)} - \proj_{V_{I}}\bb[\big]{r^{(j)}}(\Xv)
\nonumber
\\
&=
\rv^{(j)}  - \Kv \Ev_{I}^\top \Kv_{I, I}^{-1} \Ev_I \rv^{(j)}
\nonumber
\\
&=
\rv^{(j)} - \Kv_{:, I} \Kv_{I, I}^{-1} \rv^{(j)}_I,
\label{eq:text_update_of_r}
\end{align}
where we recall that $\Ev_{I}$ denotes the rows of the identity matrix indexed by $I$.

\textbf{Store $s^{(j)}$ by RKHS Bases.} 
We prove by induction that $s^{(j)} \in V_{[n]}$ for every $j$ and thus can be written as a linear combination $\sum_{i=1}^{n} w_i^{(j)} k(\xv_i, \cdot)$ for some weight $\wv^{(j)} \in \Rb^n$.
At the $0$-th iteration, we see that $s^{(0)}$ is the zero function with the weight vector $\wv^{(0)} = \zero$.
Let $I \subseteq [n]$ be the indices selected in the $j$-th iteration.
By the induction hypothesis, we have
\begin{align*}
s^{(j + 1)} &= s^{(j)} + \proj_{V_I} \bb[\big]{r^{(j)}}
\\
&= \sum_{i=1}^{n} w_i^{(j)} k(\xv_i, \cdot) + r^{(j)}(\Xv)^\top \Ev_{I}^\top \Kv_{I, I}^{-1} \Ev_{I} k(\Xv, \cdot),
\end{align*}
where the last line gives an explicit update on $\wv$:
\begin{align*}
    \wv^{(j + 1)} = \wv^{(j)} + \Ev_I^\top \Kv_{I, I}^{-1} \rv_I^{(j)}.
\end{align*}
Recalling the property of left multiplication with $\Ev_I^\top$, only entries in $\wv$ indexed by $I$ need to be updated, while keeping the entries outside $I$ unchanged:
\begin{align}
\label{eq:text_update_of_w}
\begin{split}
    \wv^{(j + 1)}_{I} &= \wv^{(j)}_I + \Kv_{I, I}^{-1} \rv^{(j)}_I, \\
    \wv^{(j + 1)}_{[n] \setminus I} &= \wv^{(j)}_{[n] \setminus I}.
\end{split}
\end{align}

\paragraph{Summary.}
The updates \eqref{eq:text_update_of_r} and \eqref{eq:text_update_of_w} yield iterations on the residual $\rv^{(j)}$ and weight vector $\wv^{(j)}$ by simple matrix operations.
As $j \to \infty$, we have $s^{(j)}$ converges to $\proj_{V_{[n]}} (g)$.
As a result, the residual vector $\rv^{(j)}$ converges to zero and the weight vector $\wv^{(j)} \to \Kv^{-1} \bv$.
We summarize this approach in \Cref{alg:alternating_projection}.
Note that the algorithm can be adapted easily to perform multiple right-hand solves in parallel
by replacing vectors $\wv, \rv, \bv$ with matrices $\Wv, \Rv, \Bv$.

\begin{algorithm}[t]
\caption{Alternating Projection}
\label{alg:alternating_projection}
\DontPrintSemicolon
\KwIn{A batched linear system $\Kv \Wv\!=\!\Bv$}
\KwOut{The solution $\Wv^* = \Kv^{-1} \Bv$}
Initialize $\Wv = \Ov$ and $\Rv = \Bv$ \\
\For(\tcp*[f]{epoch}){$t = 1, 2, \cdots$}{
    \For(\tcp*[f]{mini-batch}){$j = 1, 2, \cdots, m$}{
        select a block $I \in \Pc$ from the partition

        $\Wv_{I} = \Wv_{I} + \Kv_{I, I}^{-1} \Rv_{I}$

        $\Rv = \Rv - \Kv_{:, I} \Kv_{I, I}^{-1} \Rv_{I}$
    }
    \uIf{$\| \Rv \| \leq \epsilon \| \Bv \|$}{
      \Return $\Wv$
    }
}
\end{algorithm}

\paragraph{Block Selection.}
Selecting which block to update is crucial for fast convergence.
The simplest block selection rules are random selection (sample $I$ uniformly from $\Pc$) and cyclic selection (the $j$-th iteration selects the $(j \mod m)$-th block), which usually converge slowly (see \Cref{fig:illustration_alternating_projection}).
A more sensible choice is selecting the block $I$ with the largest residual norm
\begin{align}
\label{eq:approximate_gauss_southwell}
    I = \argmax_{I \in \Pc} ~ \lVert \Rv_{I, :} \rVert_{\mathrm{F}}^2.
\end{align}
In the special case when $\Rv$ is an $n \times 1$ vector, the selection rule \eqref{eq:approximate_gauss_southwell} reduces to the Gauss-Southwell (GS) rule \citep{nutini2015coordinate}.
When $\Rv$ is a matrix, however, the selection rule \eqref{eq:approximate_gauss_southwell} is not the same as applying the GS rule independently in each column, which may select different blocks for different columns.
Thus, the convergence behavior of \Cref{alg:alternating_projection} on multiple right hand sides is not exactly the same as running the algorithm on each right hand side independently.

\paragraph{Cached Cholesky Factors.}
Updating $\Wv$ and $\Rv$ requires solving a linear system with the submatrix $\Kv_{I, I}$. To avoid repeatedly inverting the same matrices, we compute and cache the Cholesky factors of all principal submatrices $\{ \Kv_{I, I}: I \in \Pc \}$ once at the beginning of \Cref{alg:alternating_projection}.
Namely, we cache the Cholesky factors whenever the GP hyperparameters are updated, \ie, once per gradient computation.
To facilitate parallelism, we partition the blocks evenly so that every block has the same size $\lvert I \rvert = b$ (except for the last block) and factorize all submatrices in a single batch Cholesky call.
Caching Cholesky factors costs $\Oc(n b^2)$ time and $\Oc(nb)$ memory.

\paragraph{Complexity.}
The block selection takes $\Oc(nb)$ time.
With the cached Cholesky factors available, updating the weights $\Wv$ takes $\Oc(b^2)$ time and updating the residual $\Rv$ takes $\Oc(n b)$ time.
Each epoch runs $m = n / b$ inner iterations and thus takes $\Oc(n b + n^2)$ time in total---each epoch has the same complexity as a single CG iteration.
A more fine-grained analysis in \S\ref{sec:flops_counting} shows that each epoch has $(2 + \frac{3}{b}) n^2 + (2b + 1) n$ floating point operations (FLOPs).
Thus, for typical batch sizes $1 \ll b \ll n$, each epoch requires roughly the same $2 n^2$ FLOPs as a single CG iteration.
In the upcoming sections, we will compare the total number of CG iterations and the total number of alternating projection epochs, as a proxy of comparing FLOPs.
We note that every inner iteration in \Cref{alg:alternating_projection} has linear (in terms of $n$) time and memory complexity.
In particular, the peak memory complexity is $\Oc(nb)$.

\paragraph{Connection with Coordinate Descent.}
It can be shown that \Cref{alg:alternating_projection} produces iterates equivalent to block coordinate descent on the quadratic form (\S\ref{sec:coordinated_descent_vs_alternating_projection}).
We will exploit this connection to give a convergence rate.
While block coordinate descent is arguably more intuitive, we introduce this method as alternating projection for two reasons.
First, unlike in coordinate descent, the update rules based on alternating projection maintain the residual $\Rv$ incrementally, which enables efficient block selection rules like \eqref{eq:approximate_gauss_southwell} without re-evaluating the residual.
Ultimately, block coordinate descent has to be implemented as \Cref{alg:alternating_projection} for efficiency.
Second, alternating projection can be easily adapted to new settings.
For instance, a parallel coordinate descent algorithm was discovered via the connection with (Dykstra's) alternating projection \citep{boyle1986method,tibshirani2017dykstra} in the setting of regularized least-squares, which hints that \Cref{alg:alternating_projection} may be distributed.

\section{CONVERGENCE}
\label{sec:convergence}

\begin{figure}
\centering
\begin{subfigure}{0.48\linewidth}
    \includegraphics[width=\linewidth]{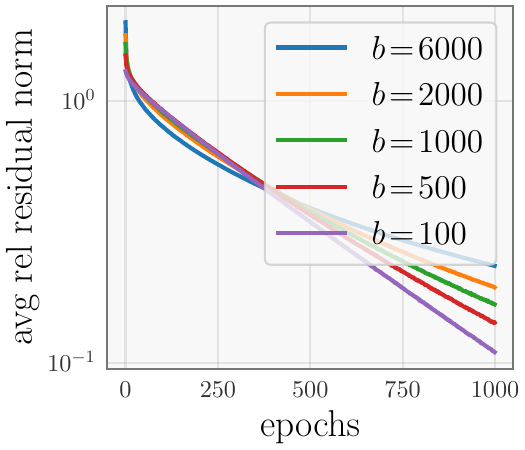}
\end{subfigure}
\begin{subfigure}{0.48\linewidth}
    \includegraphics[width=\linewidth]{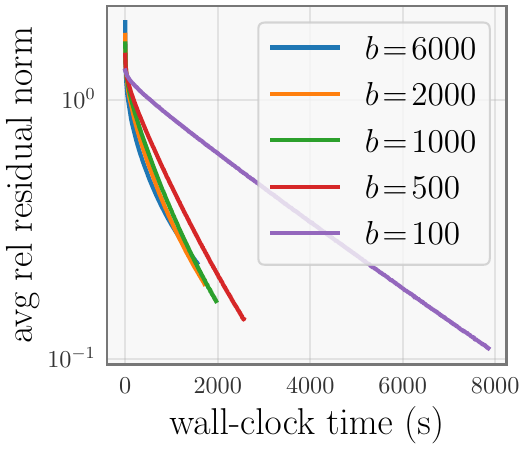}
\end{subfigure}
\caption{
Convergence of alternating projection with different batch sizes $b$ on 3droad.
\textbf{Left:}
Smaller batch sizes converge faster within the same epochs.
\textbf{Right:}
However, smaller batch sizes result in more sequential updates on the GPU and thus longer wall-clock time.
}
\label{fig:different_batch_3droad}
\end{figure}

Let $\lambda_{\max}$ and $\lambda_{\min}$ be the largest and smallest eigenvalues of $\Kv$, $\kappa = \lambda_{\max} / \lambda_{\min}$ its condition number, and define $\lambda_{\max}^\prime = \max_{I \in \Pc} \lambda_{\max} (\Kv_{I, I})$ as the maximum of the largest eigenvalues of the principal submatrices $\{K_{I, I}: I \in \Pc\}$.
By leveraging the connection with coordinate descent \citep{nutini2022let}, we can prove an explicit convergence rate for \Cref{alg:alternating_projection} when applied to a linear system with multiple right-hand sides.

\begin{restatable}{theorem}{ConvergenceGSMultipleRHS}
\label{thm:convergence-gauss-southwell}
Let $\Wv^*$ be the unique solution of the linear system \(\Kv \Wv = \Bv\) and $\Wv^{(t)}$ its approximation after $t$ epochs of \Cref{alg:alternating_projection} using the modified GS rule \eqref{eq:approximate_gauss_southwell}.
Then it holds that
\begin{align*}
    \lVert \Wv^{(t)} - \Wv^* \rVert_\Kv^2 \leq \exp\big(- t / \kappa^\prime \big) \lVert \Wv^{(0)} - \Wv^* \rVert_\Kv^2,
\end{align*}
where $\norm{\Wv - \Wv^*}_{\Kv}^2 = \tr\bb[\big]{(\Wv - \Wv^*)^\top \Kv (\Wv - \Wv^*)}$ and $\kappa^\prime = \lambda_{\max}^\prime/\lambda_{\min} \leq \kappa$.
\end{restatable}

The rate in \Cref{thm:convergence-gauss-southwell} improves over gradient descent despite only needing submatrices, for which the above holds with $\exp(- t / \kappa)$, since generally $\kappa^\prime \leq \kappa$.
For comparison, the convergence rate of (batched) CG is $4 \big((\sqrt{\kappa} - 1) / (\sqrt{\kappa} + 1)\big)^{2t} \approx 4 \exp\big(- 4t / \sqrt{\kappa}\big)$
for a sufficiently large condition number $\kappa \gg 1$.
The convergence rate of alternating projection is asymptotically faster than that of CG if $\kappa^\prime \leq \frac14 \sqrt{\kappa}$.
In general, we do not expect this condition to hold.
However, alternating projection has practical advantages despite a slower asymptotic convergence rate.
First, alternating projection has $n / b$ times more updates than CG with the same number of floating point operations.
More frequent updates may leads to more progress especially in the beginning of the optimization.
Second, alternating projection generally decreases the residual in every epoch, whereas CG residual is well-known to be non-monotonic.
Empirically, CG often increases the residual dramatically in the early stage and it takes time for CG to enter the ``linear convergence phase''. 
\jw{I think the main point here where alternating projection shines, is that it implicitly works on better conditioned matrices, which matches what we see in experiments. This is an important point to make here.}
\jw{Reference \Cref{fig:illustration_alternating_projection} (Right) somewhere and explain that random block selection only achieves the rate in \Cref{thm:convergence-gauss-southwell} in expectation.}

\Cref{fig:first-figure} demonstrates the above two points.
This figure is plotted using two checkpoints at the $50$-th epoch of GP training on the 3droad and house electric datasets respectively.
The batched linear system $\Kv \Wv = \Bv$ has $16$ right-hand sides, where $\bv_0 = \yv - \muv$ is the difference between the training labels and prior mean and $\cbb{\bv_i}_{i=1}^{15}$ are \iid stochastic trace samples.

\Cref{fig:illustration_alternating_projection} right panel compares the convergence rates of different block selection rules.
We can show that the random selection rule achieves a similar rate as \Cref{thm:convergence-gauss-southwell}, but only in expectation \citep{nesterov2012efficiency}.
In practice, the GS rule almost always converges faster than random selection.

The batch size $b$ affects the rate in \Cref{thm:convergence-gauss-southwell} through the ratio $\kappa^\prime = \lambda_{\max}^\prime/\lambda_{\min}$.
Note that the largest eigenvalue of the principal submatrix is bounded by its trace $\lambda_{\max}(\Kv_{I, I}) \leq \tr\bb{\Kv_{I, I}}$, where the trace grows linearly in $\lvert I \rvert$.
A small batch size $b = \lvert I \rvert$ is likely to have a small eigenvalue $\lambda_{\max}^\prime$ and thus a faster convergence rate (at least according to \Cref{thm:convergence-gauss-southwell}).
Indeed, as shown in \Cref{fig:different_batch_3droad}, we compare convergence rates of different batch sizes in practice.
Although small batch sizes lead to faster convergence rates, they generally have a longer running time due to more sequential updates.
Therefore, we recommend using the largest batch size possible subject to memory constraints.
In addition, we note that the convergence rate in \Cref{thm:convergence-gauss-southwell} is loose for large batch sizes $b$.
In the extreme case where $b = n$, \Cref{alg:alternating_projection} is equivalent to the Cholesky decomposition on the full kernel matrix $\Kv$ and thus converges to the exact solution in one update.
However, \Cref{thm:convergence-gauss-southwell} does not reflect that.
Hence, the convergence rate in practice may be faster than the theory predicts.

\section{EXPERIMENTS}
\label{sec:experiement}
\begin{table*}[t]
\small
\caption{
Gaussian process training on UCI benchmark datasets. Metrics are computed across multiple runs and reported with $\pm$ one standard deviation.}
\label{tab:gp_training_matern25}
\centering
\begin{tabular}{c c c c c r c}
\toprule
Dataset & Method & RMSE & NLL & CG iters/AP epochs & Training time & Speed up \\
\midrule
\multirowcell{3}{sgemm \\ $n = 241,600$ \\ $d = 14$} & CG & $ 0.048 \pm 0.000 $ & $ \bm{-1.037 \pm 0.001} $ & $551 \pm 1$ & $9.1$m $ \pm 0.0 $ & \\
& Alt. Proj. & $ \bm{0.046 \pm 0.000} $ & $ -0.999 \pm 0.001$ & $550 \pm 0$ & $12.2$m $\pm 0.2 $ & $0.7\times$ \\
\cdashline{2-7}\noalign{\vskip 0.5ex}
& SVGP & $ 0.086 \pm 0.000 $ & $ -0.934 \pm 0.003$ & NA & $14.8$m $\pm 0.1$ \\
\midrule
\multirowcell{3}{air quality \\ $n = 382,168$ \\ $d = 13$} & CG & $ \bm{0.261 \pm 0.001} $ & $ 0.143 \pm 0.004$ & $2965 \pm 19$ & $33.5$m $\pm 1.5$ \\
& Alt. Proj. & $ \bm{0.262 \pm 0.001} $ & $ \bm{0.137 \pm 0.003} $ & $550 \pm 0$ & $16.9$m $\pm 0.5$ & $2.0\times$ \\
\cdashline{2-7}\noalign{\vskip 0.5ex}
& SVGP & $ 0.363 \pm 0.003 $ & $ 0.399 \pm 0.006$ & NA & $23.4$m $\pm 0.1$ \\
\midrule
\multirowcell{3}{3droad \\ $n = 434,874$ \\ $d = 3$} & CG & $\bm{0.069 \pm 0.000}$ & $ 1.324 \pm 0.002$ & $5128 \pm 114$ & $53.2$m $\pm 2.8$ \\
& Alt. Proj. & $ 0.076 \pm 0.000 $ & $ 1.203 \pm 0.001$ & $676 \pm 1$ & $21.1$m $\pm 0.5$ & $2.5\times$ \\
\cdashline{2-7}\noalign{\vskip 0.5ex}
& SVGP & $0.327 \pm 0.002$ & $\bm{0.320 \pm 0.005}$ & NA & $26.1$m $\pm 0.1$ \\
\midrule
\multirowcell{3}{song \\ $n = 515,345$ \\ $d = 90$} & CG & $ \bm{0.747 \pm 0.002} $ & $ 1.140 \pm 0.003$ & $4431 \pm 110$ & $13.8$h $\pm 0.8$ \\ & Alt. Proj. & $ \bm{0.749 \pm 0.002} $ & $\bm{1.132 \pm 0.002}$ & $550 \pm 0$ & $2.7$h $\pm 0.1$ & $5.1\times$ \\
\cdashline{2-7}\noalign{\vskip 0.5ex}
& SVGP & $ 0.790 \pm 0.002 $ & $ 1.184 \pm 0.002 $ & NA & $0.5$h $\pm 0.0$ \\
\midrule
\multirowcell{3}{buzz \\ $n = 583,250$ \\ $d = 77$} & CG & $0.321^* \pm 0.144$ & $0.669^*\pm 1.152$ & $16726 \pm 2724$ & $31.1$h $\pm 5.4$ \\
& Alt. Proj. & $\bm{0.239 \pm 0.001}$ & $\bm{0.018 \pm 0.003}$ & $ 550 \pm 0 $ & $2.0$h $\pm 0.1$ & $15.6\times$ \\
\cdashline{2-7}\noalign{\vskip 0.5ex}
& SVGP & $ 0.259 \pm 0.002 $ & $0.066 \pm 0.006$ & NA & $0.6$h $\pm 0.0$ \\
\midrule
\multirowcell{3}{house electric \\ $n=2,049,280$ \\ $d = 11$} & CG & - & - & $\geqslant 50441$ & $\geqslant 11$d \\
& Alt. Proj. \hfill & $ \bm{0.030 \pm 0.000} $ & $ -1.148 \pm 0.001 $ & $1100 \pm 0$ & $9.8$h $\pm 0.4$ & $\geqslant 26.9\times$ \\
\cdashline{2-7}\noalign{\vskip 0.5ex}
& SVGP & $ 0.050 \pm 0.000 $ & $ \bm{-1.549 \pm 0.001} $ & NA & $2.1$h $\pm 0.0$ \\
\midrule
\multirowcell{3}{gas sensors \\ $n = 4,178,504$ \\ $d = 17$} & CG & - & - & - & 
 -  \\ 
& Alt. Proj. & $\bm{0.203}$ & $\bm{0.070}^\dagger$ & 1100 & $84.5$h & \\
\cdashline{2-7}\noalign{\vskip 0.5ex}
& SVGP & $ 0.330 \pm 0.001 $ & $ 0.339 \pm 0.003$ & NA & $8.7$h $\pm 0.03$ \\
\bottomrule
\multicolumn{7}{l}{\scriptsize *\,: At test time, CG does not reach the tolerance $\epsilon = 0.01$ after $4000$ iterations on some checkpoints.}\\
\multicolumn{7}{l}{\scriptsize -\ : CG does not finish GP training.}\\
\multicolumn{7}{l}{\scriptsize $\dagger$\,: This predictive variance is calculated using only $500$ Lanczos iterations to save time and avoid numerical instability.}
\end{tabular}
\end{table*}

We evaluate the efficacy of the alternating projections solver in a GP regression task.
Our evaluation includes a training dataset of $n = 4M$, which, to the best of our knowledge, is considerably larger than any other dataset where a GP has been applied without inducing points or employing modeling approximations.
Our implementation is available at \url{https://github.com/kayween/alternating-projection-for-gp}.

Experiments are performed on a single $24$ GB NVIDIA RTX A5000 GPUs with single precision floating point arithmetic.
All numerical algorithms and GP models are implemented in PyTorch and GPyTorch \citep{gardner2018gpytorch}.
We use the KeOps library \citep{charlier2021kernel} to implement all matrix-free numerical methods in a map-reduce fashion,
thus eliminating the need to store large $n \times n$ kernel matrices in memory.

\subsection{Main Result: GP Regression}
We first evaluate our method on large-scale Gaussian process training tasks.
We compare against GPs trained with CG, which is the predominant matrix-free GP training approach \citep{gardner2018gpytorch,wang2019exact,maddox2022low}.

\paragraph{Metrics.}
Our primary desiderata for GPs are 1) low computational costs for training and 2) generalization.
Hence, we report the following metrics for each training method:
1) the wall-clock {\bf training time},
and 2/3) the trained model's {\bf RMSE} and {\bf NLL} measured on the test set.
Additionally, for CG-trained and alternating projection-trained GPs, we report the total number of CG iterations and alternating projection epochs.

\paragraph{Datasets and Models.}
We conduct experiments on UCI regression datasets, whose statistics are shown in \Cref{tab:gp_training_matern25}.
Each dataset is split into $80\%$ training and $20\%$ test.
The labels are normalized so that they have zero mean and unit variance.
Almost all experiments are averaged over $5$ runs.
Because of resource constraints, we limit the two largest datasets---house electric and gas sensors---to 3 and 1 run respectively.

We train GPs with $\nu=2.5$ Mat\'ern kernels and a constant prior mean.
We optimize the following hyperparameters: a scalar constant for the prior mean, a $d$-dimensional kernel lengthscale, a scalar outputscale, and a scalar observational noise $\sigma^2$.
Experiments with $\nu = 1.5$ Mat\'ern kernels are deferred to \S\ref{sec:additional_experiments}.

\paragraph{MLL Optimization.}
To compute the stochastic MLL gradient \eqref{eqn:loss_grad_unbiased},
we use $l \!=\! 15$ stochastic trace samples $\zv_i$.
Thus, all matrix-free methods solve a batched linear system with $16$ right-hand sides with $\bv_0 = \yv - \muv$ and $\bv_i = \zv_i$ for $1 \leq i \leq 15$ in each training iteration.
On the first five datasets, the GPs are trained by $50$ iterations of Adam with a step size $0.1$.
On house electric and gas sensors, the GPs are trained by $100$ iterations of Adam with a step size $0.1$.

\paragraph{Alternating Projection Details.}
As discussed in \S\ref{sec:convergence}, a large batch size is preferred empirically.
We use the largest batch size that we can fit on a $24$ GB GPU.
The batch sizes $b$ are set as:
$6000$ on sgemm, air quality and 3droad;
$4000$ on song and buzz;
$1000$ on house electric;
$500$ on gas sensors.
We use the sequential partition $\Pc$: the data points from $(j - 1) b + 1$ to $j b$ belong to the $j$-th block $I_j$ for $j = 1, 2, \cdots n / b$.

The maximum CG iterations and the maximum number of alternating projection epoch is set to $1000$.
Following GPyTorch's CG stopping criteria, we terminate the alternating projection solves after
(a) the average relative residual norm is strictly smaller than the tolerance $\epsilon = 1$
or (b) $1000$ total epochs, whichever comes first.
In addition, we ensure that at least $11$ epochs of alternating projections have been run before termination (again following GPyTorch).
We define the average relative residual norm as
$\frac{1}{l + 1} \sum_{i = 0}^{l} \lVert \rv_i \rVert / {\lVert \bv_i \rVert}$
when there are $l + 1$ right hand sides $(\begin{matrix}\bv_0 & \bv_1 & \cdots & \bv_l\end{matrix})$.

\paragraph{CG Details.}
We use GPyTorch's implementation of CG,
which uses the same stopping criteria as our alternating projection implementation.
Following \citet{wang2019exact,wenger2022preconditioning}, we use a pivoted Cholesky preconditioner of size $500$ on all datasets except:
house electric uses a size $300$ and gas sensors uses a size $150$ due to GPU memory overflow.

\vspace{-.1em}

\paragraph{Prediction.}
At test time, the predictive mean is computed by the same iterative method used for training, \ie, CG for the CG-trained GP, alternating projection for the AP-trained GP.
A limitation of our method is that it does not easily result in a cache for predictive variances.
Therefore, we use $1000$ Lanczos iterations as in \cite{pleiss2018constant,wang2019exact}.

\vspace{-.1em}

\paragraph{Results on $10^5 < n < 10^6$ Datasets.}
In \Cref{tab:gp_training_matern25}, we compare the predictive performance and the training speed of CG-based versus alternating projection-based GPs.
Both training procedures produce GPs with similar RMSE and NLL.
We conjecture that this similarity occurs because both approaches solve linear systems up to the same tolerance, and thus find similar hyperparameters.
One exception is the buzz dataset: CG struggles to converge during training, resulting in considerably worse RMSE and NLL.

The primary difference between the two methods is training time.
Alternating projection-based training is up to $27\times$ faster than CG.
The only exception is sgemm, which seems to be a well-conditioned dataset since CG converges quickly.

\begin{figure}[t]
\centering
\begin{subfigure}{0.48\linewidth}
    \includegraphics[width=\linewidth]{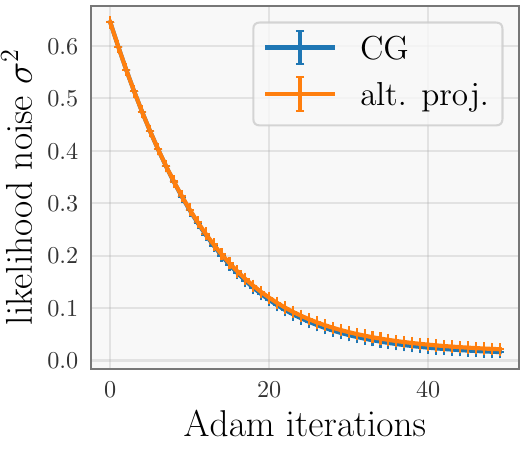}
\end{subfigure}
\begin{subfigure}{0.48\linewidth}
    \includegraphics[width=\linewidth]{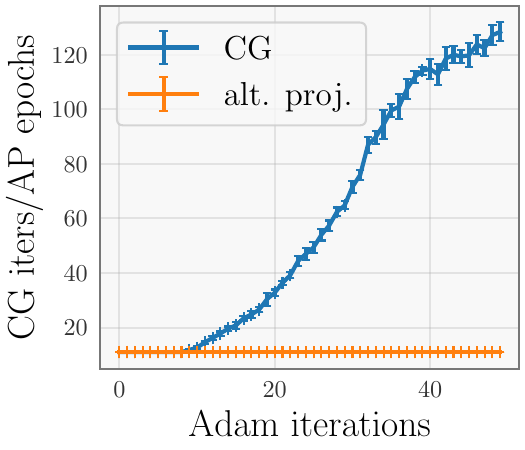}
\end{subfigure}
\caption{
GP training with Adam on air quality.
\textbf{Left:} As the likelihood noise $\sigma^2$ decreases in training, the kernel matrix $\Kv$ gets more ill-conditioned.
\textbf{Right:}
The $y$-axis is the number of iterations for CG and the number of epochs for alternating projection.
CG is sensitive to this increased ill-conditioning, while alternating projections is robust.
}
\label{fig:iterations_as_training_goes}
\end{figure}

For reference, we also report the training/test performance of SVGPs with $1024$ inducing points (see \S\ref{sec:additional_experiments} for experimental design details).
GPs trained by alternating projection achieve substantially lower RMSE and comparable NLL compared with SVGP.
We do note that SVGPs have lower NLL on 3droad and house electric,
which we suspect is a limitation of the Lanczos predictive variance estimates.
SVGP's predictive variances can be computed exactly and do not make use of the Lanczos estimator, while the predictive variances of CG/AP-trained GPs are approximated by $1000$ Lanczos iterations.
Indeed, in \S\ref{sec:additional_experiments} we find that the NLL gap shrinks as we increase the rank of the Lanczos variance estimator,
suggesting that this gap is not a fundamental limitation of the alternating projections training methodology.

\paragraph{Results on $n\geq10^6$ Datasets.}
Previous attempts to train GPs using iterative methods on datasets with $n\geq 10^6$ examples have used a large noise constraint $\sigma^2 \geq 0.1$ to improve the conditioning of the kernel matrix \citep[\eg,][]{wang2019exact,maddox2022low}.
Since alternating projection is much less conditioning-sensitive than CG (as we will see soon in \S\ref{sec:results_conditioning}), for the first time, we are able to train the GP with a much smaller noise constraint $\sigma^2 \geq 10^{-4}$, the default in GPyTorch for the Gaussian likelihood.\footnote{GPyTorch likelihood setting \url{https://rb.gy/fv41w}}
Removing the large noise constraint in hyperparameter optimization yields much better predictive performance: the RMSE $0.030$ is significantly lower than what can be achieved with high-noise constraint models (\cf \S\ref{sec:additional_experiments}).

We additionally train a GP on the gas sensors dataset with $4$ million data points.
To the best of our knowledge, this is the largest dataset trained on using GPs without the use of inducing points or other modeling approximations.
CG training appears to be intractable on such a large dataset, requiring over a month.
In contrast, alternating projection finishes training in $84.5$ hours.

\subsection{Effect of Kernel Matrix Conditioning}
\label{sec:results_conditioning}
We observe empirically that alternating projection is less sensitive to ill-conditioning than CG.
\Cref{fig:iterations_as_training_goes} shows this phenomenon, which depicts training on the $n \approx 400K$ air quality dataset.
Over the course of training, the observation noise parameter $\sigma^2$ decreases for both methods,
resulting in increasingly ill-conditioned kernel matrices (as $\lambda_{\min}(\Kv) \approx \sigma^2)$.
At the end of training, when $\sigma^2 \approx 0.01$,
CG requires over $120$ iterations to converge---$10\times$ as many iterations as the beginning of training.
In contrast, alternating projection consistently converges in $11$ iterations despite the decreasing noise and increasing condition number.
See more datasets in \S\ref{sec:additional_experiments}.

\subsection{Alternating Projection at Test Time}
\label{sec:test-time}
\begin{figure}[t]
\centering
\begin{subfigure}{0.48\linewidth}
    \includegraphics[width=\linewidth]{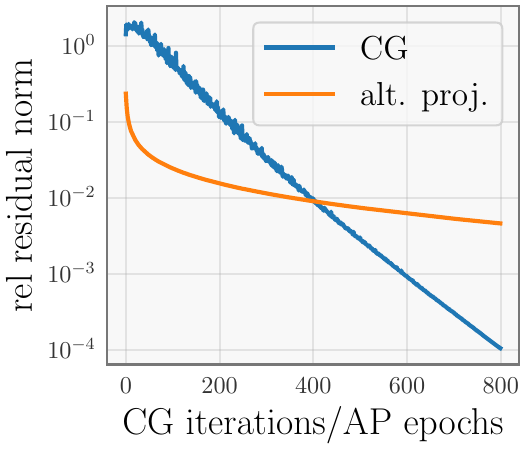}
    \caption{air quality}
\end{subfigure}
\begin{subfigure}{0.48\linewidth}
    \includegraphics[width=\linewidth]{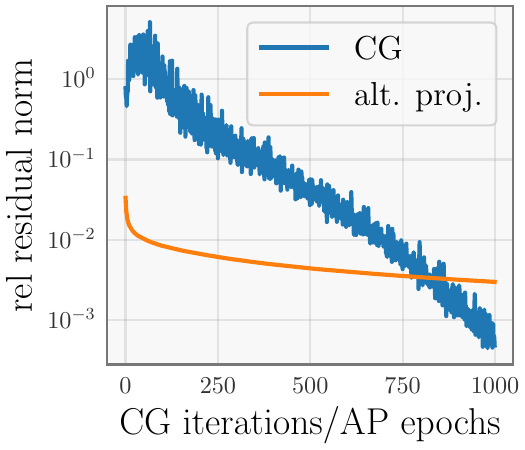}
    \caption{buzz}
\end{subfigure}
\caption{
Running CG and alternating projection on test-time solves $\Kv^{-1} (\yv - \muv)$.
The $x$-axis is the number of iterations for CG and the number of epochs for alternating projection.
\textbf{Left:}
CG has a faster asymptotic convergence rate, but CG does not reach the test-time tolerance $\epsilon = 0.01$ much faster.
\textbf{Right:}
Alternating projection reaches the tolerance $\epsilon = 0.01$ faster despite its slower asymptotic convergence rate.
}
\label{fig:test_time_solve}
\end{figure}

\begin{table*}[t]
\small
\centering
\caption{
Compute the predictive mean of the \emph{same} GP using CG and alternating projection.
}
\label{tab:predictive_mean_cg_vs_altproj}
\begin{tabular}{c c c c c r r c}
\toprule
\multirowcell{2}{Dataset} & \multicolumn{2}{c}{RMSE} & \multicolumn{2}{c}{CG iterations/AP epochs} & \multicolumn{2}{c}{Time} & \multirowcell{2}{Speed up} \\
\cmidrule(lr){2-3} \cmidrule(lr){4-5} \cmidrule(lr){6-7}
& CG & Alt. Proj. &  CG & Alt. Proj. & CG & Alt. Proj. & \\
\midrule
sgemm & $0.046 \pm 0.000$ & $0.046 \pm 0.000$ & $95 \pm 2$ & $17 \pm 0$ & $35.0$s $\pm 1.1$ & $13.5$s $\pm 0.3$ & 0.4$\times$ \\
air quality & $0.256 \pm 0.001$ & $0.256 \pm 0.001$ & $374 \pm 33$ & $388 \pm 85$ & $2.8$s $\pm 0.3$ & $3.6$s $\pm 0.8$ & 0.7$\times$ \\
3droad & $0.076 \pm 0.000$ & $0.076 \pm 0.000$ & $1586 \pm 31$ & $1720 \pm 79$ & $5.8$m $\pm 0.4$ & $9.6$m $\pm 0.6$ & $0.6\times$ \\
song & $0.749 \pm 0.002$ & $0.749 \pm 0.001$ & $211 \pm 7$ & $86 \pm 6$ & $38.1$m $\pm 0.7$ & $16.4$m $\pm 1.0$ & $2.3\times$ \\
buzz & $0.241 \pm 0.001$ & $0.239 \pm 0.001$ & $579 \pm 72$ & $41 \pm 10$ & $1.2$h $\pm 0.6$ & $4.4$m  $\pm 1.2$ & $17.2\times$ \\
house electric & $0.032 \pm 0.000$ & $0.030 \pm 0.000$ & $2111 \pm 375$ & $24 \pm 0$ & $5.6$h $\pm 0.6$ & $4.7$m $\pm 0.2$ & $72.3\times$ \\
gas sensors & $0.203$ & $0.203$ & $560$ & $13$ & $16.1$h & $27.7$m & $34.9\times$ \\
\bottomrule
\end{tabular}
\end{table*}

Any linear solver can be used to compute the posterior mean on the test data, by solving the linear system $\Kv^{-1} (\yv - \muv)$.
We explore alternating projection at test time, shown in \Cref{tab:predictive_mean_cg_vs_altproj}.
With a test-time tolerance $\epsilon = 0.01$, the posterior mean computed by alternating projection is practically the same as CG: the RMSE computed by both methods are the same up to the $3$rd digit after the decimal point in most cases.
While alternating projection is slightly slower on medium-size datasets such air quality and 3droad, we observe strong speed up on larger datasets.
In particular, alternating projection computes the posterior mean $17.2\times$ faster in wall-clock time than CG on buzz, and computes the posterior mean on house electric in $5$ min---$72\times$ faster than CG.

\Cref{fig:test_time_solve} plots the convergence CG and alternating projection at test time.
Even though CG has faster asymptotic convergence rates, alternating projection reaches the test-time tolerance $\epsilon = 0.01$ faster.
Note that CG does find high precision solutions quicker, \eg, $\epsilon = 10^{-4}$, but they are seldom necessary for GP predictions \citep{wang2019exact,maddox2022low}.

\section{RELATED WORK}
The early usage of conjugate gradients (CG) in GP training and inference dates back at least to \citet{gibbs1997efficient}.
Later, \citet{yang2004efficient,shen2005fast} proposed methods speeding up CG by approximate matrix-vector multiplications.
More recently, CG has been revisited by \citet{davies2015effective,cutajar2016preconditioning} on larger datasets with various preconditioners.
Then, a series of work \citep{pleiss2018constant,gardner2018gpytorch,wang2019exact,artemev2021tighter} and software packages such as GPyTorch \citep{gardner2018gpytorch} and GPflow \citep{matthews2017gpflow} have popularized CG for GP training and inference.

Alternating projection \citep{von1949rings} is a general algorithm finding a point in the intersection of convex sets.
The method presented in \S\ref{sec:method} is a special case in the reproducing kernel Hilbert space, and has been applied to radial basis function interpolation \citep{beatson2001fast,wendland2004scattered}.
The method turns our to be equivalent to block coordinate descent and we provide a self-contained explanation in \S\ref{sec:coordinated_descent_vs_alternating_projection}.
An early work applying coordinate descent to GPs with greedy block selection is done by \citet{bo2012greedy}.
However, the greedy block selection rule is not parallelizable on modern hardware like GPUs due to the inherent sequential nature of greedy selection.

\citet{lin2023sampling} have recently proposed an approximate GP posterior sampling method.
Their method uses stochastic gradient descent (SGD) to minimize an approximate objective based on random Fourier features and inducing points approximation.
SGD generally converges sublinearly due to stochastic noise, and its step size requires manual tuning.
Though, SGD could have cheaper per iteration cost independent of the data size $n$.
In contrast, alternating projection enjoys linear convergence with no parameters to tune, and thus may be easier to use in practice.
It would be interesting to apply our method in sampling as well to compare with \citet{lin2023sampling}.

\section{CONCLUSION}
In this work, we propose an alternating projection method with a linear convergence rate for solving dense kernel linear systems and applied it to GP training and inference.
The method quickly reaches commonly used tolerances faster than CG, requires only linear time per iteration, and is more robust to ill-conditioning.
Experiments on several large-scale benchmark datasets show that the method achieves a speed-up of up to $27\times$ over CG-based training and of up to $72\times$ over CG-based inference.
We are able to train and evaluate GPs on millions of data points without artificially inflating the observation noise for stability, leading to increased predictive performance.
In particular, this includes a dataset with $4$ million data points, to the best of our knowledge, the largest dataset reported in the literature so far without inducing point approximation.

\subsection*{Acknowledgements}
The authors thank the anonymous reviewers for their helpful comments.
KW, HJ and JRG are supported by NSF award IIS-2145644.
JW was supported by the Gatsby Charitable Foundation (GAT3708), the Simons Foundation (542963), the NSF AI Institute for Artificial and Natural Intelligence (ARNI: NSF DBI 2229929) and the Kavli Foundation.

\clearpage

\bibliography{ref}

\clearpage

\section*{Checklist}

 \begin{enumerate}

 \item For all models and algorithms presented, check if you include:
 \begin{enumerate}
\item A clear description of the mathematical setting, assumptions, algorithm, and/or model.

Yes.
\item An analysis of the properties and complexity (time, space, sample size) of any algorithm.

Yes.
\item (Optional) Anonymized source code, with specification of all dependencies, including external libraries.

Yes.
 \end{enumerate}

 \item For any theoretical claim, check if you include:
 \begin{enumerate}
\item Statements of the full set of assumptions of all theoretical results.

Yes.
\item Complete proofs of all theoretical results.

Yes.
\item Clear explanations of any assumptions.

Yes.     
 \end{enumerate}

 \item For all figures and tables that present empirical results, check if you include:
 \begin{enumerate}
\item The code, data, and instructions needed to reproduce the main experimental results (either in the supplemental material or as a URL).

Yes.

\item All the training details (e.g., data splits, hyperparameters, how they were chosen).

Yes.

\item A clear definition of the specific measure or statistics and error bars (e.g., with respect to the random seed after running experiments multiple times).

Yes.

\item A description of the computing infrastructure used. (e.g., type of GPUs, internal cluster, or cloud provider).

Yes.
 \end{enumerate}

\item If you are using existing assets (e.g., code, data, models) or curating/releasing new assets, check if you include:
\begin{enumerate}
\item Citations of the creator if your work uses existing assets.

Yes.

\item The license information of the assets, if applicable.

Not applicable.

\item New assets either in the supplemental material or as a URL, if applicable.

Not applicable.

\item Information about consent from data providers/curators.

Not applicable.

\item Discussion of sensible content if applicable, e.g., personally identifiable information or offensive content.

Not applicable.
\end{enumerate}

\item If you used crowdsourcing or conducted research with human subjects, check if you include:
\begin{enumerate}
\item The full text of instructions given to participants and screenshots.

Not applicable.

\item Descriptions of potential participant risks, with links to Institutional Review Board (IRB) approvals if applicable.

Not applicable.

\item The estimated hourly wage paid to participants and the total amount spent on participant compensation.

Not applicable.

\end{enumerate}

 \end{enumerate}

\onecolumn
\appendix

\aistatstitle{\papertitle: \\[10pt]Supplementary Material}

\startcontents[sections]
\printcontents[sections]{l}{1}{\setcounter{tocdepth}{3}}
\vfill

\clearpage

\section{Von Neumann's Alternating Projection}
\label{sec:von_neumann_alternating_projection}
This section shows that the method presented in \S\ref{sec:method} is indeed a special case of von Neumann's alternting projection \citep{von1949rings}.
Let $g \in \Hc$ be a function that interpolates the data, \ie, $g(\Xv) = \bv$.
Write $g$ as an orthogonal decomposition
\begin{align*}
    g = \proj_{V_{[n]}} (g) + \proj_{V_{[n]}^\perp} (g),
\end{align*}
where $V_{[n]}^\perp$ is the orthogonal complement of $V_{[n]}$.
Thus, computing $\proj_{V_{[n]}} (g)$ reduces to computing the projection to the orthogonal complement $V_{[n]}^\perp$.
Write the orthogonal complement in the form
\begin{align*}
    V_{[n]}^\perp = \bigcap_{I \in \Pc} V_I^\perp = \bigcap_{I \in \Pc} \cbb{f \in \Hc: f(\xv_i) = 0 \forall i \in I},
\end{align*}
which is an intersection of $n$ convex sets.
Starting from $f^{(0)} = g$, the $j$-th iteration of alternating projection selects a block $I$ and computes a projection
\begin{align}
\label{eq:von_neumann_alternating_projection}
    f^{(j + 1)} = \proj_{V_I^\perp} \bb[\big]{f^{(j)}}
\end{align}
As $j \to \infty$, we have $f^{(j)} \to \proj_{V_{[n]}}^\perp (g)$ and $g - f^{(j)} \to \proj_{V_{[n]}} (g)$.
Recall the identity $\proj_{V_I^\perp} (f) = f - \proj_{V_I} (f)$.
Thus, \eqref{eq:von_neumann_alternating_projection} implies
\begin{align*}
    f^{(j + 1)} = f^{(j)} - \proj_{V_I} \bb[\big]{f^{(j)}},
\end{align*}
which is exactly the same as the update rule \eqref{eq:update_r} on the residual $r^{(j)}$.
As a result, $g - f^{(j)}$ is exactly the same as $s^{(j)}$ in the update \eqref{eq:update_s}.

\section{Connection between Coordinate Descent and Alternating Projection}
\label{sec:coordinated_descent_vs_alternating_projection}

This section presents the connection between \Cref{alg:alternating_projection} and coordinate descent, as shown in \Cref{alg:coordinate_descent}.

\begin{algorithm}[h]
\caption{Block Coordinate Descent}
\label{alg:coordinate_descent}
\DontPrintSemicolon
\KwIn{A kernel linear system $\Kv \Wv = \Bv$}
\KwOut{The solution $\Kv^{-1} \Bv$}
Initialize $\Wv = \Ov$ \\
\For(\tcp*[f]{epoch}){$i = 1, 2, \cdots$}{
    \For(\tcp*[f]{mini-batch}){$j = 1, 2, \cdots, m$}{
        select a block $I \in \{I_1, I_2, \cdots, I_m\}$ \\
        \label{line:cholesky_solve}
        $\Wv_{I} = \Kv_{I, I}^{-1} \big(\Bv_{I} - \Kv_{I, \neg I} \Wv_{\neg I}\big)$ \\
    }
    \uIf{converged}{
      \Return $\Wv$
    }
}
\end{algorithm}

Observe that the minimizer of the quadratic objective
\begin{equation}
\label{eqn:quadratic-objective}
h(\Wv) = \frac{1}{2} \tr\bb[\big]{\Wv^\top \Kv \Wv} - \tr\bb[\big]{\Bv^\top \Wv} 
\end{equation}
is exactly the solution $\Kv^{-1} \Bv$ of the linear system $\Kv \Wv = \Bv$.

Given a partition of indices $\{I_1, I_2, \cdots, I_{m}\}$ where $I_i \cap I_j = \emptyset$ for all $i \neq j$ and $\cup_{i=1}^{m} I_i = [n]$, coordinate descent minimizes \eqref{eqn:quadratic-objective} by minimizing over a subset of variables $\Wv_I = \Wv_{I, :}$ in each iteration.
Taking the derivative \wrt the subblock $\Wv_I$, we have
\begin{align*}
[\nabla h(\Wv)]_I & = \Kv_I \Wv - \Bv_I \\
& = 
\bb*{\begin{matrix}
    \Kv_{I, I} & \Kv_{I, \neg I}
\end{matrix}}
\bb*{\begin{matrix}
    \Wv_{I} \\
    \Wv_{\neg I}
\end{matrix}}
- \Bv_I,
\end{align*}
where the second line splits $\Kv_I$ and $\Wv$ into two blocks.
The index $\neg I = [n] \setminus I$ denotes the complement of $I$.
Setting the derivative to zero gives the following update
\begin{align*}
    \Wv_I^{(j + 1)} = \Kv_{I, I}^{-1} \big(\Bv - \Kv_{I, \neg I} \Wv_{\neg I}^{(j)}\big)
\end{align*}
which minimizes \eqref{eqn:quadratic-objective} over $\Wv_I$ exactly.
The full algorithm of coordinate descent is shown in \Cref{alg:coordinate_descent}.

The following lemma shows the $\Rv$ matrix in \Cref{alg:alternating_projection} is indeed the residual of the linear system.
This lemma will be useful in proving the equivalence between \Cref{alg:alternating_projection} and \Cref{alg:coordinate_descent}.
\begin{lemma}
\label{lma:residual_equality}
Let $\Rv^{(j)}$ and $\Wv^{(j)}$ be the residual and weight after $j$ updates of \Cref{alg:alternating_projection}.
Then, we have
\begin{align*}
    \Rv^{(j)} = \Bv - \Kv \Wv^{(j)}.
\end{align*}
\end{lemma}

\begin{proof}
The proof is based on an induction on the number of updates $j$ (the number of inner loops).
At the initialization $j = 0$, the equality holds trivially.
Suppose after the $j$-th update we have $\Rv^{(j)} = \Bv - \Kv \Wv^{(j)}$.
All we need to do is to verify this equality in the case of $j + 1$ by direct calculation:
\begin{align*}
    \Bv - \Kv \Wv^{(j + 1)} & = \Bv - \Kv \big(\Wv^{(j)} + \Ev_I^\top \Kv_{I, I}^{-1} \Ev_I \Rv^{(j)}\big) \\
    & = \Rv^{(j)} - \Kv \Ev_I^\top \Kv_{I, I}^{-1} \Ev_I \Rv^{(j)} \\
    & = \Rv^{(j + 1)}
\end{align*}
where the first line uses the update rule \eqref{eq:text_update_of_w} of $\Wv^{(j)}$ and the last line uses the update rule \eqref{eq:text_update_of_r} of $\Rv^{(j)}$.
\end{proof}

With \Cref{lma:residual_equality}, we are ready to show the equivalence between \Cref{alg:alternating_projection} and \Cref{alg:coordinate_descent}.

\begin{lemma}
Let $\Wv^{(j)}$ be the weight produced by \Cref{alg:alternating_projection} after $j$ updates.
Them, we have
\begin{align*}
    \Wv_I^{(j + 1)} & = \Kv_{I, I}^{-1} \big(\Bv_{I} - \Kv_{I, \neg I} \Wv_{\neg I}^{(j)}\big), \\
    \Wv_{\neg I}^{(j + 1)} & = \Wv_{\neg I}^{(j)},
\end{align*}
where $\neg I = [n] \setminus I$.
Thus, \Cref{alg:alternating_projection} produces the same iterates as \Cref{alg:coordinate_descent}.
\end{lemma}

\begin{proof}
Recalling the update rule \eqref{eq:text_update_of_w}, we have
\begin{align*}
    \Wv^{(j + 1)} = \Wv^{(j)} + \Ev_I^\top \Kv_{I, I}^{-1} \Ev_I \Rv^{(j)}.
\end{align*}
Recalling the property of left multiplication with $\Ev_I^\top$, entries outside $I$ are unchanged and thus $\Wv_{\neg I}^{(j + 1)} = \Wv_{\neg I}^{(j)}$.

On the other hand, entries indexed by $I$ satisfy $\Wv_I^{(j+1)} = \Wv_I^{(j+1)} + \Kv_{I, I}^{-1} \Ev_I \Rv^{(j)}$.
Plug in $\Rv^{(j)} = \Bv - \Kv \Wv^{(j)}$ by \Cref{lma:residual_equality} and thus we have
\begin{align*}
    \Wv_I^{(j + 1)} & = \Wv_I^{(j)} + \Kv_{I, I}^{-1} \Ev_I \big(\Bv - \Kv \Wv^{(j)}\big) \\
    & = \Wv_I^{(j)} + \Kv_{I, I}^{-1} \big(\Bv_I - \Kv_{I} \Wv^{(j)}\big) \\
    & = \Wv_I^{(j)} + \Kv_{I, I}^{-1} \big(\Bv_I - \Kv_{I, I} \Wv_I^{(j)} - \Kv_{I, \neg I} \Wv_{\neg I}^{(j)}\big) \\
    & = \Kv_{I, I}^{-1} \big(\Bv_I - \Kv_{I, \neg I} \Wv_{\neg I}^{(j)}\big)
\end{align*}
where the second line uses the definition of $\Ev_I$; the third line split the matrix $\Kv_I$ into blocks $\Kv_I = \big(\begin{matrix}\Kv_{I, I} & \Kv_{I, \neg I}\end{matrix}\big)$; the last line is straightforward algebra.
\end{proof}

\section{Proofs}
\label{sec:proofs}

\begin{restatable}{lemma}{}
\label{lem:h-strongly-convex}
The quadratic objective function \eqref{eqn:quadratic-objective} satisfies the Polyak-Łojasiewicz (PL) inequality
\begin{align*}
    \frac12 \| \nabla h(\Wv) \|_{\mathrm{F}}^2 \geq \lambda_{\min} (h(\Wv) - h(\Wv^*)),
\end{align*}
where $\lambda_{\min} > 0$ is the smallest eigenvalue of $\Kv$.
\end{restatable}

\begin{proof}
If $\Wv$ has only a single column this follows directly from the strong convexity of the quadratic function.
When $\Wv$ has multiple columns, $h$ is a separable function across each column.
Therefore, $h$ is also $\lambda_{\min}$ strongly convex which implies the PL inequality.
\end{proof}

\begin{restatable}{lemma}{}
\label{lem:repr-weights-Knorm-equals-objective}
For $h(\Wv)$ as in \eqref{eqn:quadratic-objective}, it holds that $h(\Wv) - h(\Wv^*) = \frac12 \| \Wv - \Wv^* \|_{\Kv}^2$.
\end{restatable}
\begin{proof}
Plugging \(\Bv = \Kv \Wv^*\) into the expression of $h$, straightforward algebra gives
\begin{align*}
h(\Wv) - h(\Wv^*) & = \frac12 \langle\Wv, \Kv \Wv\rangle - \langle \Bv, \Wv\rangle - \frac12 \langle\Wv^*, \Kv \Wv^*\rangle + \langle\Bv, \Wv^*\rangle \\
& = \frac12 \langle\Wv, \Kv \Wv\rangle - \langle \Kv \Wv^*, \Wv\rangle - \frac12 \langle\Wv^*, \Kv \Wv^*\rangle + \langle\Kv \Wv^*, \Wv^*\rangle \\
& = \frac12 \langle\Wv, \Kv \Wv\rangle - \langle \Kv \Wv^*, \Wv\rangle + \frac12 \langle\Wv^*, \Kv \Wv^*\rangle \\
& = \frac12 \| \Wv - \Wv^* \|_{\Kv}^2.
\end{align*}
\end{proof}

\ConvergenceGSMultipleRHS*

\begin{proof}
By straightforward algebra, the improvement on the objective $h$ as in \eqref{eqn:quadratic-objective} after the $j$-th update is
\begin{align*}
    h(\Wv^{(j + 1)}) - h(\Wv^{(j)}) = - \frac12 \norm[\big]{\Rv_{I, :}^{(j)}}_{\Kv_{I, I}^{-1}}^2.
\end{align*}
For any residual $\Rv$ matrix, note the following inequality
\begin{align}
\label{eqn:bound_residual_using_gs_rule}
    \| \Rv \|_{\mathrm{F}}^2 & = \sum_{I \in \Pc} \| \Rv_{I, :} \|_{\mathrm{F}}^2 \leq \lvert \Pc \rvert \cdot \max_{I \in \Pc} \| \Rv_{I, :} \|_{\mathrm{F}}^2 = m \cdot \max_{I \in \Pc} \| \Rv_{I, :} \|_{\mathrm{F}}^2.
\end{align}
Thus, the improvement on the objective $h$ is bounded by
\begin{align*}
    h(\Wv^{(j + 1)}) - h(\Wv^{(j)}) & \leq - \frac{1}{2 \lambda_{\max}^\prime} \| \Rv_{I, :}^{(j)} \|_{\mathrm{F}}^2 \\
    & \leq - \frac{1}{2 m \lambda_{\max}^\prime} \| \Rv^{(j)} \|_{\mathrm{F}}^2
\end{align*}
where the first inequality is because $\frac{1}{\lambda_{\max}^\prime}$ is the smallest eigenvalue of $\Kv_{I, I}$;
the second inequality is due to the Gauss-Southwell selection rule and \eqref{eqn:bound_residual_using_gs_rule}.
Subtract $h^* = h(\Wv^{*})$ from both sides.
Then, we have
\begin{align*}
    h(\Wv^{(j + 1)}) - h^* & = h(\Wv^{(j)}) - h^* - \frac{1}{2 m \lambda_{\max}^\prime} \| \Rv^{(j)} \|_{\mathrm{F}}^2 \\
    & \leq \big(1 - \frac{\lambda_{\min}}{m \lambda_{\max}^\prime}\big) \big(h(\Wv^{(j)}) - h^*\big) \\
    & \leq \big(1 - \frac{1}{m \kappa^\prime}\big) \big(h(\Wv^{(j)}) - h^*\big)
\end{align*}
where the second line uses $\Rv^{(j)} = \Bv - \Kv \Wv^{(j)} = - \nabla h(\Wv^{(j)})$ by \Cref{lma:residual_equality} and the PL inequality by \Cref{lem:h-strongly-convex}.
Using the inequality $(1 - x)^{t} \leq \exp(-tx)$, we obtain a convergence rate in the number of updates $j$:
\begin{align*}
    h(\Wv^{(j + 1)}) - h^* \leq \exp\Big(-\frac{j}{m \kappa^\prime}\Big) \big(h(\Wv^{(0)}) - h^*\big).
\end{align*}
Since each epoch has $m$ updates, the convergence rate in the number of epochs $t$ is
\begin{align*}
    h(\Wv^{(t + 1)}) - h^* \leq \exp\Big(-\frac{t}{\kappa^\prime}\Big) \big(h(\Wv^{(0)}) - h^*\big).
\end{align*}
By \Cref{lem:repr-weights-Knorm-equals-objective}, the left and right hand sides can be written as \(\| \Wv^{(t)} - \Wv^* \|_{\Kv}^2\) and \(\| \Wv^{(0)} - \Wv^* \|_{\Kv}^2\) respectively, which concludes the proof.

\end{proof}

\section{Descriptions of the UCI Datasets in the Experiments}
This section lists the relevant information of the datasets with citations.
The datasets used in the papers are sgemm GPU \citep{sgemmgpu}, air quality \citep{beijing_air_quality}, 3droad \citep{3droad}, song \citep{song}, buzz \citep{yang2015la}, house electric \citep{houseelectric}, and gas sensors \citep{gassensors}.
All of them are downloaded from the UCI machine learning repository \citep{uci}.

\section{Additional Experiments}
\label{sec:additional_experiments}
This section presents more experimental details and additional experiments.

\subsection{Further Experimental Details}

\paragraph{GP Training.}
All Gaussian processes, including stochastic variational Gaussian processes, use an observation noise constraint $\sigma^2 \geq 10^{-4}$, which is the default in GPyTorch.
For the stochastic trace estimation \eqref{eqn:loss_grad_unbiased}, we use $\ell = 15$ random probe vectors.
For CG, the probe vectors are sampled from $\Nc(\zero, \Pv)$, where $\Pv$ is the pivoted Cholesky preconditioner.
Again, these settings are default in GPyTorch.
For alternating projection, the probe vectors are sampled from a Rademacher distribution.

\paragraph{Preconditioning.}
CG uses the pivoted Cholesky preconditioner both in training and test.
During training, the preconditioner size is $500$ on sgemm, air quality, 3droad, song and buzz; $300$ on house electric; $150$ on gas sensors.
We decrease the preconditioner size on house electric and gas sensors due to GPU memory overflow.
During test, the preconditioner size is $500$ on sgemm, air quality, 3droad, song, buzz and house electric; $300$ on gas sensors.
Again, we decrese the preconditioner size on gas sensors due to GPU memory flow.
See \Cref{tab:preconditioner-size-batch-size}.

\paragraph{Alternating Projection Batch Size.}
The batch sizes during training and test are shown in \Cref{tab:preconditioner-size-batch-size}.

\begin{table}[b]
\centering
\small
\caption{Preconditioner sizes and batch sizes during training and test.}
\label{tab:preconditioner-size-batch-size}
\begin{tabular}{c c c c c c c c c}
\toprule
method & train/test & sgemm & air quality & 3droad & song & buzz & house electric & gas sensors
\\
\midrule
\multirowcell{2}{CG preconditioner size}
& train & $500$ & $500$ & $500$ & $500$ & $500$ & $300$ & $150$
\\
& test & $500$ & $500$ & $500$ & $500$ & $500$ & $500$ & $300$
\\
\midrule
\multirowcell{2}{alt. proj. batch size} & train & $6000$ & $6000$ & $6000$ & $4000$ & $4000$ & $1000$ & $500$
\\
& test & $6000$ & $6000$ & $6000$ & $4000$ & $4000$ & $1000$ & $500$
\\
\bottomrule
\end{tabular}
\end{table}

\paragraph{SVGP Training.}
All SVGPs use $1024$ inducing points and a batch size of $4096$.
On the first six datasets, SVGPs are trained with $50$ epochs of Adam with a step size $0.01$ and another $150$ epochs of Adam with a step size $0.001$.
On gas sensors, we train the SVGP with $50$ epochs of Adam with a step size $0.01$ followed by $350$ epochs of Adam with a step size $0.001$.

\paragraph{Other Experimental Settings.}
The right panel of \Cref{fig:illustration_alternating_projection} is produced on with an alternating projection-trained GP on air quality with batch size $1000$.
The linear system solved in the figure is $\Kv^{-1} \yv$ (without subtracting the prior mean $\muv$).
\Cref{fig:different_batch_3droad} is plotted with an alternating projection-trained GP on 3droad.
The linear system in the figure is $\Kv^{-1} (\begin{matrix}\yv & \zv_1 & \zv_2 & \cdots & \zv_{15}\end{matrix})$ where $\zv_i$ are sampled from a standard Gaussian distribution.

\subsection{GP Training on House Electric with Large Noise Constraint $\sigma^2 \geq 0.1$}
We compare Gaussian processes on house electric trained with two different noise constraints $\sigma^2 \geq 0.1$ and $\sigma^2 \geq 10^{-4}$, as shown \Cref{tab:houseelectric_large_noise_vs_small_noise}.
We observe significant improvements on both RMSE and NLL when the noise is smaller.
In particular, the GP trained with small noise constraint $\sigma^2 \geq 10^{-4}$ has $40\%$ smaller RMSE and significantly smaller NLL.
This indicates that artificially inflating the observation noise $\sigma^2$, while making the kernel matrix well-conditioned, ultimately hurts the predictive performance.

With alternating projection, training the GP with a small noise constraint $\sigma^2 \geq 10^{-4}$ is as fast as the GP with a large noise constraint $\sigma^2 \geq 10^{-1}$.
The RMSE and NLL are computed with the same settings as the main paper.

\begin{table}[h]
\footnotesize
\centering
\caption{Comparison of GP training on the house electric dataset with large noise constraint $\sigma^2 \geq 0.1$ and small noise constraint $\sigma^2 \geq 10^{-4}$.}
\label{tab:houseelectric_large_noise_vs_small_noise}
\begin{tabular}{c c c c c r}
\toprule
Dataset & Method & RMSE & NLL & CG iterations/AP epochs & Time \\
\midrule
\multirowcell{3}{house electric \\ $n=2,049,280$ \\ $d = 11$} & CG \hfill ($\sigma^2 \geq 10^{-1}$) & $0.050 \pm 0.000$ & $-0.196 \pm 0.000$ & $1200 \pm 8$ & $9.6$h $\pm 0.6$ \\
& Alt. Proj. \hfill ($\sigma^2 \geq 10^{-1}$) & $0.053 \pm 0.000$  & $-0.197 \pm 0.000$ & $1100 \pm 0$ & $9.8$h $\pm 0.4 $ \\
& Alt. Proj. \hfill ($\sigma^2 \geq 10^{-4}$) & $ \bm{0.030 \pm 0.000} $ & $ \bm{-1.148 \pm 0.001} $ & $1100 \pm 0$ & $9.8$h $\pm 0.4$ \\
\bottomrule
\end{tabular}
\end{table}

\subsection{CG Iterations During Training}
\begin{figure*}[h]
\centering
\begin{subfigure}{0.24\linewidth}
    \begin{subfigure}{\linewidth}
        \includegraphics[width=\linewidth]{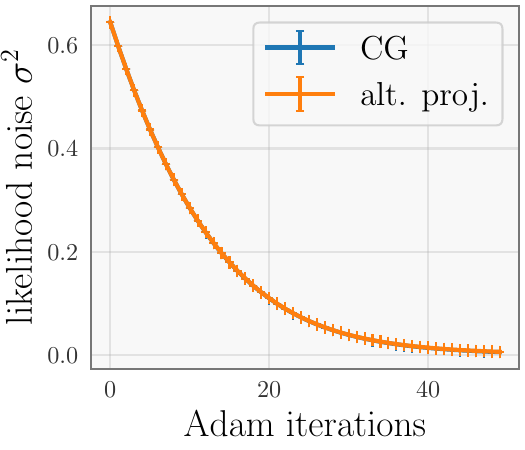}
    \end{subfigure}
    \begin{subfigure}{\linewidth}
        \includegraphics[width=\linewidth]{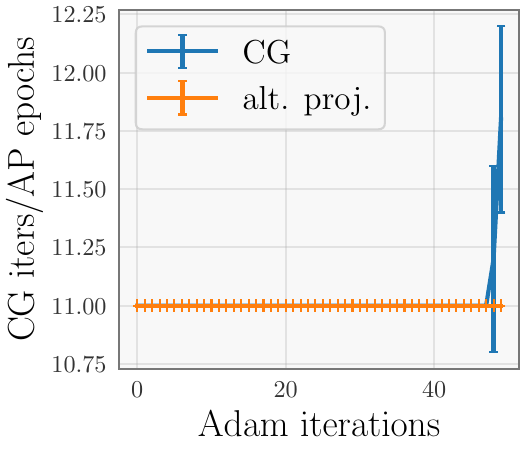}
    \end{subfigure}
\caption{sgemm GPU}
\end{subfigure}
\begin{subfigure}{0.24\linewidth}
    \begin{subfigure}{\linewidth}
        \includegraphics[width=\linewidth]{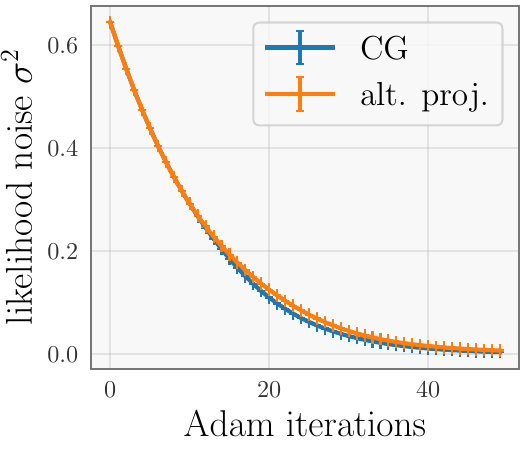}
    \end{subfigure}
    \begin{subfigure}{\linewidth}
        \includegraphics[width=\linewidth]{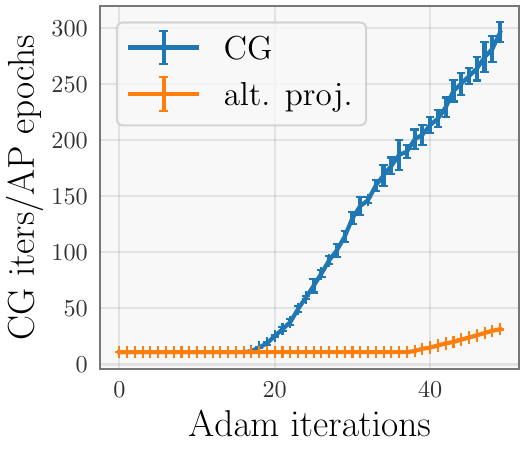}
    \end{subfigure}
    \caption{3droad}
\end{subfigure}
\begin{subfigure}{0.24\linewidth}
    \begin{subfigure}{\linewidth}
        \includegraphics[width=\linewidth]{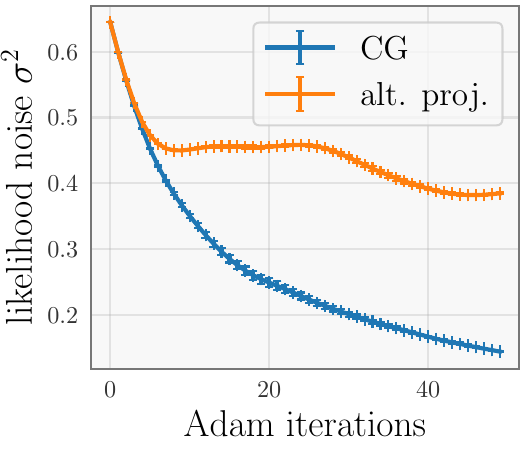}
    \end{subfigure}
    \begin{subfigure}{\linewidth}
        \includegraphics[width=\linewidth]{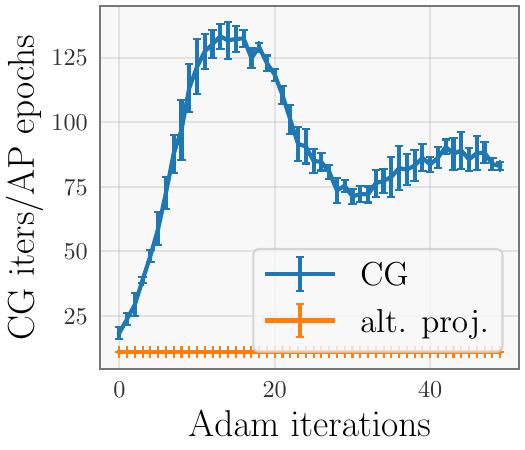}
    \end{subfigure}
    \caption{song}
\end{subfigure}
\begin{subfigure}{0.24\linewidth}
    \begin{subfigure}{\linewidth}
        \includegraphics[width=\linewidth]{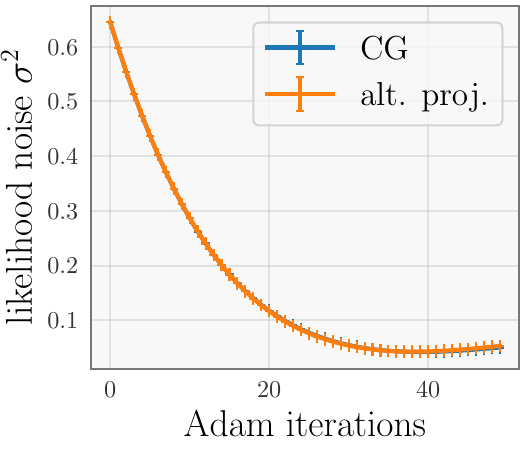}
    \end{subfigure}
    \begin{subfigure}{\linewidth}
        \includegraphics[width=\linewidth]{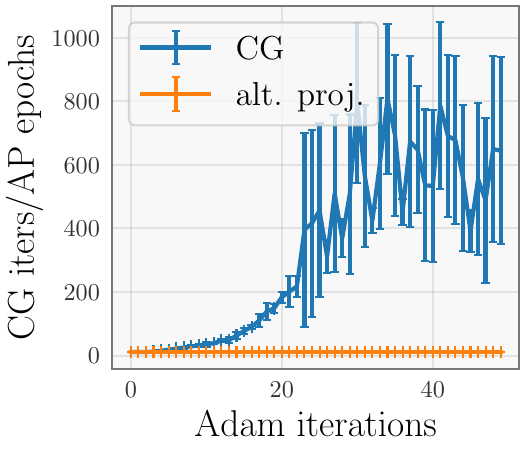}
    \end{subfigure}
    \caption{buzz}
\end{subfigure}
\caption{
The observation noise $\sigma^2$ and the number of CG iterations/alternating projection epochs during training.
\textbf{Top:} The observation noisea $\sigma^2$ decreases as the training goes.
\textbf{Bottom:} CG takes more iterations to converge as the observation noise decreases during training.
However, alternating projection is less sensitive to the decrease of observation noise.
}
\label{fig:noise_and_flops_vs_adam_iterations}
\end{figure*}

\Cref{fig:iterations_as_training_goes} in the main paper is produced on air quality.
This section presents figures on more datasets, as shown in \Cref{fig:noise_and_flops_vs_adam_iterations}.
We observe a similar phenomenon: as the noise decreases during training, the number of CG iterations increases; in contrast, alternating projection converges steadily.

\subsection{Increasing Lanczos Iterations Improves NLL}
\begin{figure}[h]
\centering
\begin{subfigure}{0.32\linewidth}
    \includegraphics[width=\linewidth]{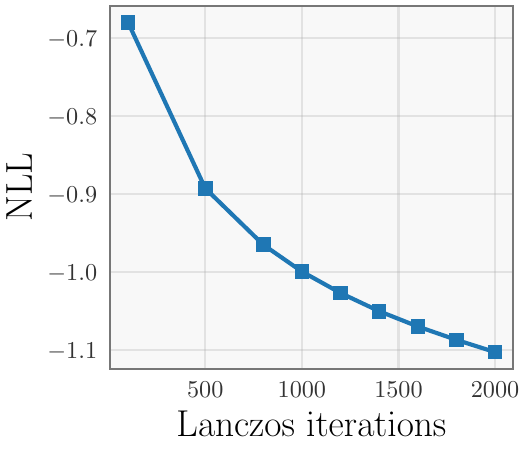}
\caption{SGEMM}
\end{subfigure}
\begin{subfigure}{0.32\linewidth}
    \includegraphics[width=\linewidth]{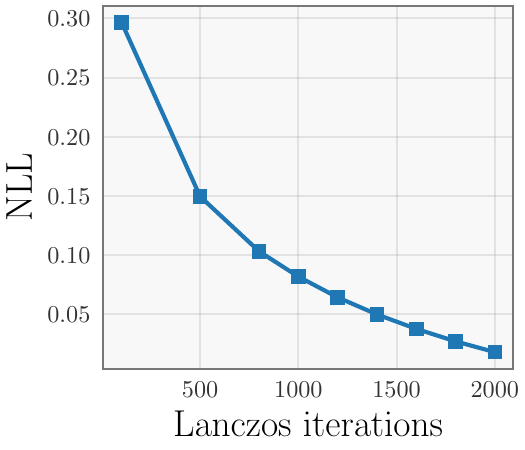}
\caption{air quality}
\end{subfigure}
\begin{subfigure}{0.32\linewidth}
    \includegraphics[width=\linewidth]{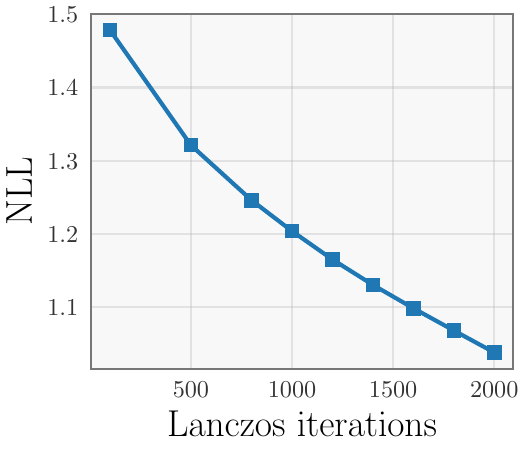}
\caption{3droad}
\end{subfigure}

\begin{subfigure}{0.32\linewidth}
    \includegraphics[width=\linewidth]{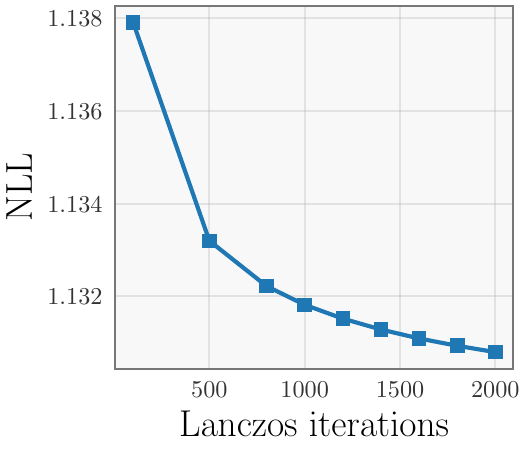}
\caption{song}
\end{subfigure}
\begin{subfigure}{0.32\linewidth}
    \includegraphics[width=\linewidth]{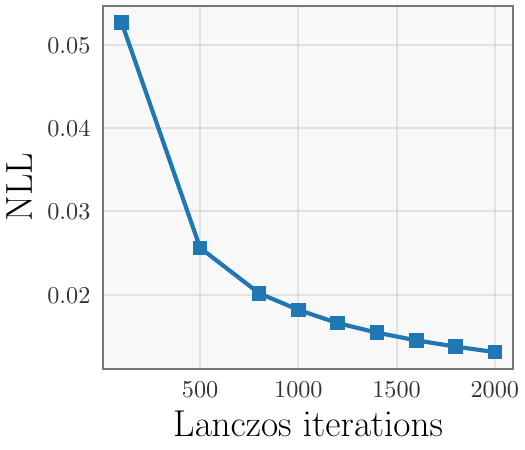}
\caption{buzz}
\end{subfigure}
\begin{subfigure}{0.32\linewidth}
    \includegraphics[width=\linewidth]{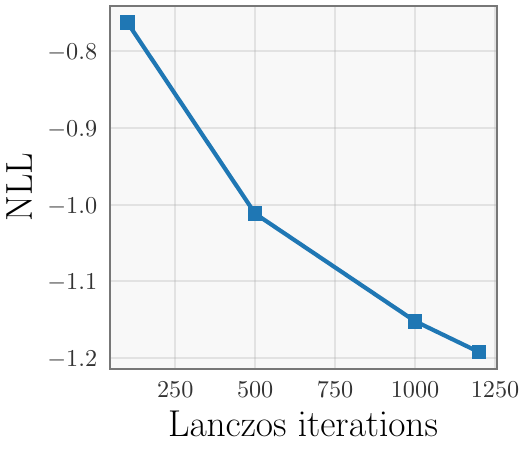}
\caption{house electric}
\end{subfigure}
\caption{
Test negative log likelihood (NLL) \vs the number of Lanczos iterations.
Empirically, the test NLL decreases as the number of Lanczos iterations increases on all datasets.
}
\label{fig:nll_vs_lanczos}
\end{figure}

In the experiments, we use $1000$ Lanczos iterations to compute the predictive variance and the test negative log likelihood (NLL).
This section investigates the relation between test NLL and the Lanczos iterations, as shown in \Cref{fig:nll_vs_lanczos}.
We empirically observe that increasing the Lanczos iterations always decreases the test NLL.
This suggests that the true NLL of the GPs may be even lower than what is reported in \Cref{tab:gp_training_matern25}.

\subsection{Training Gaussian Processes with Mat\'ern $\nu = 1.5$}

Lastly, we report results using Mat\'ern $\nu = 1.5$.
The experimental settings are exactly the same as Mat\'ern $\nu = 2.5$ GPs.
We observe a similar phenomenon: while CG-trained GPs and alternating projection-trained GPs have similar RMSE and NLL, alternating projection achieves $1.4\times$ to $27.2\times$ speed up against CG.

\begin{table*}[h]
\small
\caption{
Gaussian process training on UCI benchmark datasets with Mat\'ern $\nu = 1.5$.
Metrics are computed across multiple runs and reported with $\pm$ one standard deviation.
}
\label{tab:gp_training_matern25}
\centering
\begin{tabular}{c c c c c r c}
\toprule
Dataset & Method & RMSE & NLL & CG iters/AP epochs & Training time & Speed up \\
\midrule
\multirowcell{3}{SGEMM \\ $n = 241,600$ \\ $d = 14$} & CG & $\bm{0.048 \pm 0.000}$ & $\bm{-1.071 \pm 0.001}$ & $550 \pm 0$ & $8.9$m $\pm 0.2$ \\ 
& Alt. Proj. & $\bm{0.048 \pm 0.000}$ & $-1.060 \pm 0.001$ & $550 \pm 0$ & $12.1$m $\pm 0.2$ & $0.7\times$ \\
\cdashline{2-7}\noalign{\vskip 0.5ex}
& SVGP & $0.085 \pm 0.000$ & $-0.932 \pm 0.001$ & NA & $18.3$m $\pm 0.1$ & \\ 
\midrule
\multirowcell{3}{air quality \\ $n = 382,168$ \\ $d = 13$} & CG & $\bm{0.227 \pm 0.002}$ & $0.131 \pm 0.003$ & $1825 \pm 26$ & $22.5$m $\pm 1.2$ \\
& Alt. Proj. & $0.253 \pm 0.001$ & $\bm{0.033 \pm 0.002}$ & $550 \pm 0$ & $16.1$m $\pm 0.5$ & $1.4\times$ \\
\cdashline{2-7}\noalign{\vskip 0.5ex}
& SVGP &  $0.358 \pm 0.002$ & $0.387 \pm 0.005$ & NA & $28.8$m $\pm 0.1$ \\ 
\midrule
\multirowcell{3}{3droad \\ $n = 434,874$ \\ $d = 3$} & CG & $\bm{0.065 \pm 0.001}$ & $1.062 \pm 0.003$ & $6086 \pm 142$ & $44.4$m $\pm 2.2$ \\ 
& Alt. Proj. & $0.069 \pm 0.001$ & $0.896 \pm 0.002$ & $572 \pm 1$ & $16.5$m $\pm 0.3$ & $2.7\times$ \\ 
\cdashline{2-7}\noalign{\vskip 0.5ex}
& SVGP & $0.319 \pm 0.002$ & $\bm{0.294 \pm 0.007}$ & NA & $32.4$m $\pm 0.1$ \\ 
\midrule
\multirowcell{3}{song \\ $n = 515,345$ \\ $d = 90$} & CG & $\bm{0.743 \pm 0.001}$ & $ 1.135 \pm 0.003$ & $4393 \pm 159$ & $13.7$h $\pm 0.6$ \\
& Alt. Proj. & $\bm{0.746 \pm 0.002}$ & $\bm{1.129 \pm 0.002}$ & $550 \pm 0$ & $2.6$h $\pm 0.0$ & $5.3\times$ \\
\cdashline{2-7}\noalign{\vskip 0.5ex}
& SVGP & $0.790 \pm 0.002$ & $1.184 \pm 0.002$ & NA & $0.6$h $\pm 0.0$ \\ 
\midrule
\multirowcell{3}{buzz \\ $n = 583,250$ \\ $d = 77$} & CG & $\bm{0.238 \pm 0.000}$ & $0.027 \pm 0.002$ & $13608 \pm 2299$ & $25.4$h $\pm 4.7$ \\ 
& Alt. Proj. & $\bm{0.238 \pm 0.001}$ & $\bm{0.002 \pm 0.004}$ & $550 \pm 0$ & $1.9$h $\pm 0.1$ & $13.4\times$ \\
\cdashline{2-7}\noalign{\vskip 0.5ex}
& SVGP & $0.255 \pm 0.002$ & $0.049 \pm 0.009$ & NA & $0.7$h $\pm 0.0$ \\ 
\midrule
\multirowcell{3}{house electric \\ $n=2,049,280$ \\ $d = 11$} & CG & - & - & - & $\geqslant 11$d & \\ 
& Alt. Proj. & $\bm{0.029 \pm 0.000}$ & $-1.321 \pm 0.000$ & $1100 \pm 0$ & $9.7$h $\pm 0.1$ & $\geqslant 27.2\times$ \\
\cdashline{2-7}\noalign{\vskip 0.5ex}
& SVGP & $0.048 \pm 0.000$ & $\bm{-1.580 \pm 0.003}$ & NA & $2.6$h $\pm 0.0$ \\
\midrule
\multirowcell{3}{gas sensors \\ $n = 4,178,504$ \\ $d = 17$} & CG & - & - & - & - \\
& Alt. Proj. & $\bm{0.201}$ & $\bm{0.245}^\dagger$ & $1100$ & $42$h$^*$ \\ 
\cdashline{2-7}\noalign{\vskip 0.5ex}
& SVGP & $0.311 \pm 0.002$ & $0.286 \pm 0.004$ & NA & $10.6$h $\pm 0.1$ \\
\bottomrule
\multicolumn{7}{l}{\scriptsize $\dagger$\,: This predictive variance is calculated using only $500$ Lanczos iterations to save time and avoid numerical instability.} \\
\multicolumn{7}{l}{\scriptsize *\,: Time measured on a A100 GPU.}
\end{tabular}
\end{table*}

\section{FLOPs in \Cref{alg:alternating_projection}}
\label{sec:flops_counting}
The following table gives floating point operations (FLOPs) and memory complexity of \Cref{alg:alternating_projection}.
There is no hidden constant in the leading term.
Throughout, we assume $l \ll n$ and $1 \ll b \ll n$.
Note that the peak memory consumption is $2nb$.
We use this to estimate the largest batch $b$ that fits in a GPU.

\begin{table}[h]
\centering
\caption{FLOP Counting in \Cref{alg:alternating_projection}.}
\label{tab:flops_count}
\begin{tabular}{c c c}
\toprule
Operation & FLOPs & Memory \\
\midrule
Cache Cholesky decomposition of $\{\Kv_{I, I}: I \in \Pc\}$ & $\frac13 n b^2$ & $nb$ \\
\midrule
GS rule $I = \argmax_{I \in \Pc} ~ \lVert \Rv_{I, :} \rVert_{\mathrm{F}}^2$ & $2nl$ & - \\
\midrule
$\Wv_{I} = \Wv_{I} + \Kv_{I, I}^{-1} \Rv_{I}$ & $(b^2 + b) l$ & - \\
\midrule
$\Rv = \Rv - \Kv_{:, I} \Kv_{I, I}^{-1} \Rv_{I}$ & $(b^2 + 2nb + n) l$ & $nb$ \\
\midrule
total FLOPs of a single epoch & $\big((2 + \frac3b) n^2 + (2b + 1) n\big) l $ & $2nb$ \\
\bottomrule
\end{tabular}
\end{table}

\section{Additional Discussions}
The alternating projection method presented in this paper is not easy to parallel on multiple GPUs.
Indeed, the update for each block is sequential.
When multiple GPUs are available, CG might be more beneficial as explored by \citet{wang2019exact}.
Another limitation of alternating projection is that it does not yield an estimate of the marginal log-likelihood (MLL).
Therefore, one cannot monitor the convergence progress by plotting the MLL.
A workaround is to instead monitor the observation noise $\sigma^2$.
Typically, the observation noise $\sigma^2$ diminishes during training, and a small update in $\sigma^2$ is usually a good indication of convergence.

\citet{artemev2021tighter} utilize CG to construct a better variational lower bound for variational GPs \citep{titsias2009variational}.
Different from the stochastic variational GP  \citep{hensman2013gaussian}, this method cannot be trained by mini-batch stochastic optimization, since they plug in the closed-form solution of the variational distribution.
Interestingly, they show that warming up CG for the linear solve $\Kv^{-1} \yv$ yield a significant speed-up.
This trick might be useful in CG-based and alternating projection-based training as well.

\end{document}